\documentclass[10pt,twocolumn,letterpaper]{article}

\PassOptionsToPackage{pagebackref,breaklinks,colorlinks,allcolors=wacvblue}{hyperref}

\usepackage{wacv}              % To produce the CAMERA-READY version
\usepackage[table, dvipsnames]{xcolor}
\usepackage{pifont}
\usepackage{graphicx}
\usepackage{booktabs}
\usepackage{multirow}
\usepackage{adjustbox}
\usepackage{subcaption}
\usepackage{array}
\usepackage[normalem]{ulem}
\newcolumntype{R}[2]{%
    >{\adjustbox{angle=#1,lap=1.3\width-(#2)}\bgroup}%
    l%
    <{\egroup}%
}

\newcommand{\ra}[1]{\renewcommand{\arraystretch}{#1}}

\usepackage[accsupp]{axessibility}  % Improves PDF readability for those with disabilities.

\makeatletter
\renewcommand{\paragraph}{%
  \@startsection
    {paragraph}% name
    {4}% level
    {\z@}% indent
    {1ex plus .2ex minus .1ex}% before skip
    {-1em}% after skip (negative => run-in heading)
    {\normalfont\normalsize\bfseries}%
}
\makeatother

\usepackage{orcidlink}

\definecolor{turquoise}{rgb}{0.20, 0.70, 0.65}
\def\pmixplus{PillarMix\textsuperscript{\smash+}\xspace}
\definecolor{better}{rgb}{0.19, 0.55, 0.91}
\definecolor{worse}{rgb}{0.82, 0.1, 0.26}
\newcommand{\cmark}{\textcolor{better}{\ding{51}}}
\newcommand{\xmark}{\textcolor{worse}{\ding{55}}}

\newcommand{\renaud}[1]{\textcolor{Black}{#1}}

\definecolor{MyYellow}{RGB}{160, 160, 0}

\let\renaudRoundOne\relax
\let\spyrosRoundOne\relax
\let\alexRoundOne\relax
\let\nerminRoundOne\relax

\definecolor{wacvblue}{rgb}{0.21,0.49,0.74}
\def\wacvPaperID{774 (was 1597)} % *** Enter the WACV Paper ID here
\def\confName{WACV}
\def\confYear{2027}

\title{\renaud{VaViT:} Vanilla ViT for Automotive Point Cloud Semantic Segmentation} 
\author{
   Gilles Puy\textsuperscript{1}\thanks{Equal contribution.} \quad
   Nermin Samet\textsuperscript{1}\footnotemark[1] \quad
   Alexandre Boulch\textsuperscript{1} \\
   Spyros Gidaris\textsuperscript{1} \quad
   Tuan-Hung Vu\textsuperscript{1,3} \quad   
   Renaud Marlet\textsuperscript{1,2}
   \vspace{0.3cm} \\
   \textsuperscript{1}valeo.ai, France \quad
   \textsuperscript{2}LIGM, CNRS, Univ Gustave Eiffel, ENPC, IP Paris, France \quad \textsuperscript{3}CAIR, VinUni \\
}

\begin{document}

\maketitle

\begin{abstract}
  Plain Transformers have become the de-facto architecture for processing text, audio, image, and video, offering a unified backbone for multimodal learning. However, state-of-the-art architectures for point cloud semantic segmentation remain dominated by U-Nets architectures where convolutions are interleaved with local or windowed attentions. In this work, we show how to effectively leverage vanilla, non-hierarchical ViTs for the segmentation of large-scale automotive lidar scenes. We bridge the performance gap thanks to a carefully designed tokenizer, a lightweight decoder segmentation head, and tailored data augmentations. Our approach, VaViT, for Vanilla ViT, matches or exceeds the performance of state-of-the-art methods while maintaining the simplicity of ViT architecture. We provide extensive evaluations on nuScenes, SemanticKITTI, and Waymo Open Dataset to validate the efficiency of our method. Code and pre-trained models are available at \url{https://github.com/valeoai/VaViT}.
  
\end{abstract}

% ===============================================
%
% ===============================================
\section{Introduction}
\label{sec:intro}

Transformers have become the dominant architecture across modalities, powering state-of-the-art \renaudRoundOne{(SOTA)} systems in image \cite{dosovitskiy2021image}, language \cite{vaswani2017attention}, audio \cite{gong2021ast}, and video \cite{arnab2021vivit} processing. Their uniform design, based on self-attention and feed-forward blocks, enables a shared modeling paradigm and greatly facilitates multimodal learning \cite{xu2023multimodal}. Vision Transformers (ViTs) \cite{dosovitskiy2021image}, in particular, have demonstrated remarkable scaling properties: while Convolutional Neural Networks (CNNs) benefit from strong inductive biases and perform well in limited-data regimes, ViTs exhibit superior power-law scaling and continue to improve as model size and dataset scale increase \cite{dosovitskiy2021image, zhai2022scaling, dehghani2023scaling}. This ability to learn global dependencies without relying on handcrafted architectural priors makes the plain Transformer an appealing candidate for a unified backbone across perception tasks.

But this architectural simplicity has not translated to point cloud semantic segmentation. % Current state-of-the-art methods depart significantly from vanilla Transformers.
\renaud{Most methods labeled as ``Transformers'' actually depart significantly from va\-nil\-la ViTs: they use local forms of attention, and rely on hierarchical architectures to address long-range dependencies.}

A first line of work, exemplified by PTv3 \cite{ptv3} and LitePT \cite{litept}, adopts hybrid U-Net-style architectures \cite{ronneberger2015unet} where sparse 3D convolutions \renaud{(which limits in practice target platforms for inference)} 
are interleaved with local self attention, moving away from the plain Transformer design. % paradigm. 
Empirical evidence from LitePT actually shows that most of the “heavy lifting” is performed by the convolutional layers, questioning if they could be replaced by generic global self-attention.
Such hybridization prevents a clean unification with Transformer %-based 
architectures used with other modalities, % images, audio, or text, 
and limits the benefits of architectural standardization.

\renaud{Other approaches, including RangeViT \cite{rangevit} and RangeFormer \cite{rangeformer}, are inspired by image ViTs and operates on 2D range-view projections of lidar data. While RangeViT uses a standard ViT backbone, RangeFormer relies on a hierarchical composition of transformer blocks. Although computationally convenient, the egocentric 2D range-view projection alters the spatial 3D structure, underperforming methods that reason directly on 3D points \cite{ptv3,litept}. In contrast, Bird’s Eye View (BEV) projection respects much more the (almost) 2.5D nature of lidar scans \cite{waffleiron}.}

In this work, we ask the %% a simple 
question: can a plain, non hierarchical Vision Transformer operate %% directly 
on 3D point clouds and achieve competitive performance for large-scale \renaudRoundOne{driving} lidar semantic segmentation? We answer 
positively by introducing VaViT, a point cloud segmentation framework that %% processes raw 3D points and 
leverages a \underline{va}nilla \underline{ViT} \renaudRoundOne{backbone with global self-attention, without hierarchical stages or sparse convolution blocks}.
The success of this design relies on three key ingredients.

First, we set a dedicated tokenization strategy. Contrary to range-view approaches that convert the entire lidar scan into a 2D image representation \renaud{and rely on 2D-based embeddings}, we extract point-wise embeddings %% directly 
in 3D space and project them onto a coarse BEV %% Bird’s Eye View (BEV) 
grid, \renaudRoundOne{as in PointPillars \cite{lang2019pointpillars},} to form tokens.
Despite this projection \renaud{(not ``full 3D'')}, we show that a properly-designed tokenization stem preserves enough geometric information for downstream semantic segmentation. 
A plain ViT then processes the tokens\rlap.

Second, we design a lightweight decoder head that combines the global contextual features produced by the ViT with the high-resolution point embeddings computed prior to tokenization, enabling precise semantic labeling while keeping the overall architecture simple and efficient.

\spyrosRoundOne{Third, point cloud semantic segmentation models routinely \renaudRoundOne{and effectively} rely on strong data augmentations, in particular mixing-based strategies that generate composite scenes by combining portions of different scans. This is especially important in our setting where, compared to the large-scale datasets typically used for ViT pretraining, lidar segmentation benchmarks remain relatively small. In this context, we \renaudRoundOne{extend} % adapt 
PillarMix \cite{pillarmix} to our ViT framework through a simple occupancy-preserving modification, yielding consistent performance improvements.}

\spyrosRoundOne{Extensive experiments on nuScenes \cite{nuscenes}, SemanticKITTI \cite{semkitti} and Waymo Open Dataset \cite{waymo} show that VaViT matches or outperforms state-of-the-art methods for lidar semantic segmentation, while relying on a simple and unified ViT architecture. These results suggest that carefully engineered hybrid architectures may not be necessary for point cloud segmentation, and that a plain Vision Transformer is a competitive alternative for 3D semantic segmentation.}

\spyrosRoundOne{Overall, our work demonstrates that a vanilla ViT, when paired with a minimal set of standard components, can serve as a strong backbone for lidar semantic segmentation. Rather than introducing architectural modifications to the Transformer blocks, 
we identify key practical ingredients \nerminRoundOne{such as} tokenization, decoding, and training strategy, that make plain ViT backbones effective for this task.}

% ===============================================
%
% ===============================================
\section{Related work}
\label{sec:related}

\paragraph{Network architectures on points.} There are four broad types of network architectures to process point clouds: (1)~directly working on points, (2)~working on projected versions of the point clouds, (3)~working on voxelized representations, (4)~mixing several types of representations.

PointNet \cite{pointnet} and PointNet++ \cite{pointnetpp}, which work directly on points, pioneered the field and mainly consists of a repetition of shared MLPs applied pointwise. Several improvements have followed, by defining point convolution operators, e.g., \cite{kpconv,fkaconv}, leveraging networks developed for graph--structered data, e.g., \cite{dgcnn,SPGraph}, improving the training recipe \cite{pointnext}, or taking inspiration from architectures developed for images such as the MLP Mixer \cite{mlpmixer}, e.g., \cite{pointmixer,pointmlp,waffleiron}.

\renaudRoundOne{The commonly-used} % Among the second type of architectures, the most used projection is probably the
range projection \cite{rangenetpp,salsaNext,SqueezeSegV3,SCSSnet,KPRNet} %, which 
is a natural representation % to use 
for rotating lidars. Other works leverage projections in bird's-eye view (BEV) \cite{lang2019pointpillars, Polarnet,lmscnet}.

The release of fast GPU implementations for sparse convolutions has greatly facilitated work on voxel representations of large-scale point clouds \cite{minknet,spvnas}. The basic architectures were then improved by adapting the voxel shape to the structure of lidar point clouds \cite{Cylinder3d}. Self-distillation techniques was also used to improve performance \cite{PVKD,sdseg3d}. Attention mechanisms, added on top of such convolutions, also helped in reaching better scores \cite{AF2S3Net,gasn,svaseg,sphereformer}. Such convolutions are still today part of state-of-the-art architectures \cite{ptv3,litept}.
%
% Finally, several works have shown that 
It can also be advantageous to fuse some of the representations described above \cite{tornadonet,amvnet,gfnet,fusionnet,pcscnet,RPVNet}.

\paragraph{Transformer on points.}
% In the context of 
For large-scale 3D point cloud processing, 
directly using self-attention layers on all points is hardly feasible due to the large number
of points and the quadratic complexity of attention. 
To maintain computational efficiency while still capturing complex geometric structures, early methods like % the 
Point Transformer \cite{pointtransformer} apply % applied 
self-attentions locally and rely % relied 
on farthest point sampling to downsample the point cloud and % be able to 
capture large-scale structures.
Fast Point Transformer \cite{fastpointtrans} combines % proposed to combine 
local attention with a voxel-based hashing technique to reach fast inference times. % much faster inference time. 
Stratified Transformer \cite{strattrans} introduces a % then introduced a key  
strategy that samples nearby points densely and distant points sparsely, enlarging the receptive field in the attention layers while keeping a low computational cost. 
Superpoint Transformer \cite{superpointtransformer} constructs % proposed to construct
a hierarchical superpoint structure and captures % to capture 
the relationship between superpoints using a self-attention mechanism.
SphereFormer \cite{sphereformer} uses a % proposed the use of 
radial window self-attention to aggregate information from close and distant points from the sensor. 
PTv3 \cite{ptv3} transitions % transitioned 
toward high-efficiency serialization, using space-filling curves to reorder points into a 1D sequence, allowing for fast grouping via linear sequence chunks rather than % exhaustive 
spatial searches. 
Last, LitePT \cite{litept} improves % improved 
upon PTv3 with a lighter and better performing architecture that uses sparse convolutions in the early and last layers of the U-Net architecture and self-attentions in the the middle layers.

% In parallel to point approaches, 
Projection-based methods simplify % have simplified
tokenization by operating in 2D and leveraging a regular 2D grid structure. RangeViT \cite{rangevit} and RangeFormer \cite{rangeformer} use % both used 
range projection, which is tokenized % and tokenization 
by extracting small patches. FlatFormer \cite{flatformer} introduces % introduced 
a flattened window attention mechanism that tokenizes the point cloud by partitioning and flattening it into groups of equal sizes. FlatFormer was originally tested for object detection and applied on a BEV representation of the point clouds. VaViT and FlatFormer thus share the same type of input representation. However their downstream architecture is completely different: a regular ViT % for us, 
with global self-attention vs %; 
local attentions within groups constructed by alternating the sorting spatial dimension.

\begin{figure}[t] % RM: was figure*
\centering
\includegraphics[width=1.05\linewidth]{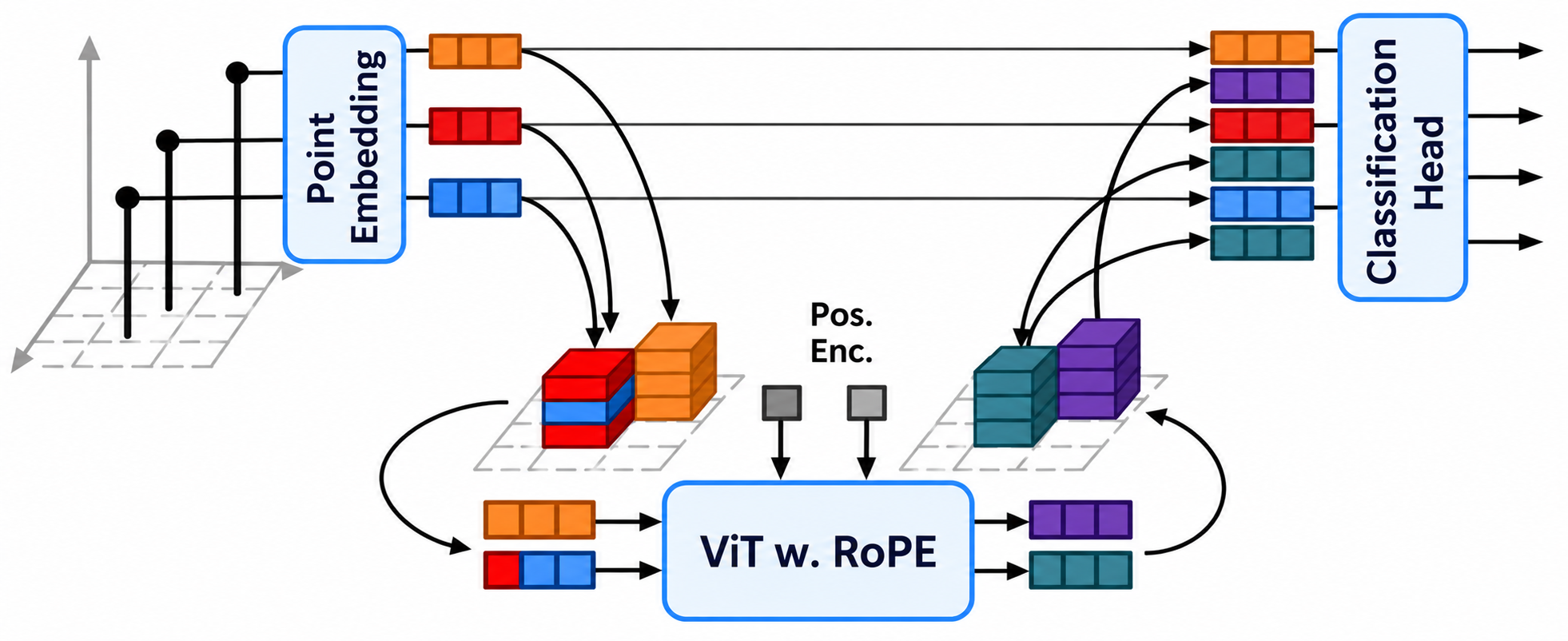}    
\caption{\textbf{Overview.} Our tokenizer aggregates point-level embeddings $\mathbf{p}_i$ into $Q$ non-empty pillar embeddings, \renaudRoundOne{serving} % which serves 
as input tokens $\mathbf{t}_q$ for a vanilla Vision Transformer. After being processed by the ViT, the $Q$ pillar tokens are redistributed
and merged with the original point embeddings, forming the final representations used for point classification. Positions on the 2D BEV space are encoded using RoPE.}
\label{fig:vavit}
\end{figure}

\paragraph{Lidar-data augmentations.}
Beyond standard augmentations such as rotation and global scaling, % it has been shown that 
mixing-based strategies can greatly 
improve % the performance of 
lidar point cloud segmentation. 
Instance CutMix \cite{RPVNet} improves the segmentation performance of rare objects by extracting such % object 
instances from a source point cloud and pasting them into a target scene. % 
However, 
simply inserting such objects in a scan breaks the lidar scan pattern. 
To better preserve this pattern, LaserMix \cite{lasermix} partitions scans into non-overlapping areas along the inclination angle and % then 
creates a new scan by mixing these partitions. 
PolarMix \cite{polarmix} does a similar operation but with non-overlapping partitions % done 
along the azimuth angle.
Last, % 
% Finally, 
complementing these partitioning schemes, PillarMix \cite{pillarmix} utilizes a pillar-based mixing approach to generate augmented training samples. We \renaudRoundOne{exploit} % propose 
an improved version, \pmixplus, 
\spyrosRoundOne{adapted} for training our ViT backbones.

% ===============================================
%
% ===============================================
\section{Method}
\label{sec:method}

Our goal is the semantic segmentation of point clouds using a plain Vision Transformer \cite{dosovitskiy2021image} operating %% directly 
on 3D points. For that, we design a pipeline that bridges the gap between irregular point cloud data and the token-based representation used by Transformers, while preserving fine geometric information for point-level prediction. 

An overview of the proposed framework is illustrated in \cref{fig:vavit}. \renaudRoundOne{Our tokenizer first}
% We first introduce a tokenizer that 
converts a point cloud into a set of pillar tokens defined in BEV, extracting expressive point-level embeddings (\cref{sec:tokenizer}). These tokens are then processed by a standard, non-hierarchical Vision Transformer that models long-range spatial dependencies across the scene (\cref{sec:method_vit}). To recover fine-grained predictions, % we design 
a lightweight segmentation head % that
redistributes the transformer tokens onto individual points and combines them with the original point embeddings (\cref{sec:method_head}). Finally, we \renaudRoundOne{adapt and tailor} % introduce 
a data augmentation strategy % tailored to 
for lidar segmentation, based on mixing pillars from multiple scans, which improves generalization when training vanilla ViTs on autonomous driving datasets (\cref{sec:method_aug}).

% ===============================================
\subsection{Tokenizer} 
\label{sec:tokenizer}

The tokenizer takes as input a list of $N$ points, each associated with low dimensional features (typically $x$, $y$, $z$ coordinates, range and intensity). These \renaudRoundOne{input} % point 
features pass in an embedding network to provide high-dimensional point embeddings. The architecture of this % embedding
network is inspired from the embedding layer used in \cite{waffleiron} and consists of a series of shared PointNet++/DGCNN-like layers~\cite{pointnet,dgcnn}. % , as described below.
The tokens given as input to the ViT are then computed in BEV: one token per square pillar of a preset side dimension.
For a given pillar, we gather the embeddings of all points falling into this pillar and the corresponding pillar token is computed by max-pooling over this set of embeddings.

Formally, let us denote by $\mathbf{p}_i^{(\ell)} \in \mathbb{R}^{F_\ell}$, % \thvc{why $N \times$?}, 
the \renaud{$F_\ell$-dimension} features of the $i^{\rm th}$ point at the $\ell^{\rm th}$ layer of the tokenizer. The tokenizer consists of a repetition of $L_{\rm emb}$ layers $f^{(\ell)}(\cdot)$ applied pointwise. The computation of features $\mathbf{p}_i^{(\ell+1)}$ is decomposed into three stages: a first stage working directly on the point feature $\mathbf{p}_i^{(\ell)}$; a second stage operating on the local representations $\mathbf{p}^{(\ell)}_j - \mathbf{p}^{(\ell)}_i$, where $j \in \mathcal{N}_i$ and $\mathcal{N}_i$ is the set of $k$-nearest neighbors (in Euclidean space) to the $i^{\rm th}$ point; a third stage combining the results of the first two.

The first stage produces an intermediate point representation obtained by applying to $\mathbf{p}^{(\ell)}_i$ a linear layer, followed by a batchnorm layer (BN) and a ReLU activation. This layer is denoted by $f_{\rm glob}^{(\ell)}(\cdot)$

For the second stage, at a local level, we start by computing all the relative representations $\tilde{\mathbf{p}}^{(\ell)}_{ij} = f_{\rm rel}^{(\ell)}(\mathbf{p}^{(\ell)}_j - \mathbf{p}^{(\ell)}_i)$ and obtain the local representation for point $i$ as 
\begin{align}
\tilde{\mathbf{p}}^{(\ell)}_{i} = \max_{j \in \mathcal{N}_i}
\left(
f_{\rm gate}^{(\ell)} ( \tilde{\mathbf{p}}^{(\ell)}_{ij} ) 
\odot 
\tilde{\mathbf{p}}^{(\ell)}_{ij}
\right),
\end{align}
where $f_{\rm rel}^{(\ell)}(\cdot)$ has the same architecture as $f_{\rm glob}^{(\ell)}(\cdot)$, i.e., a linear layer followed by BN and ReLU, $\odot$ is the channelwise multiplication, and $\max$ is applied channelwise (max-pooling). 
\renaud{In contrast to \cite{waffleiron} that directly max-pools $\smash{\tilde{\mathbf{p}}^{(\ell)}_{ij}}$, we} use a gating mechanism, with $f_{\rm gate}(\cdot)$ consisting of a linear layer followed by a sigmoid. It \renaud{selectively} discards spurious features in the set of neighbors $\mathcal{N}_i$, improving of point embedding quality for ViT downstream processing.

The third stage, which produces the final representation of point $i$ at layer $\ell + 1$, then satisfies
\begin{align}
\mathbf{p}^{(\ell + 1)}_i
= 
f_{\rm cat}^{(\ell)}
\left(
\left[
f_{\rm glob}^{(\ell)}(\mathbf{p}^{(\ell)}_i)
\; , \;
\tilde{\mathbf{p}}^{(\ell)}_{i}
\right]
\right),
\end{align}
where $f_{\rm cat}^{(\ell)}(\cdot)$ is just a linear layer and $[\cdot , \cdot]$ denotes the concatenation. This process is repeated $L_{\rm emb}$ times to obtain $\mathbf{p}^{(L_{\rm emb})}_i$. In our design, $F_{\ell} = F_{\ell+1}$ except for $F_0$ and $F_{L_{\rm emb}}$ which matches the size of the input point features and the inner dimension of the ViT, respectively.

The initial tokens $\mathbf{t}_q^{(0)}$, $q = 1, \ldots, Q$, provided as input to the ViT, are obtained by dividing the scene into regular non-overlapping square pillars of side length~$s$. For each \emph{non-empty} pillar~$q$, the representations $\mathbf{p}^{(L_{\rm emb})}_i$ of all points falling into the square are gathered \renaudRoundOne{and max-pooled} 
to \renaudRoundOne{produce} % obtain 
a pillar embedding $\mathbf{t}_q^{(0)}$ used initial token. 
Note that \renaudRoundOne{these token as computed}
only for occupied pillars.  
\renaudRoundOne{The} number of tokens $Q$ \renaudRoundOne{may thus vary} % varies 
per point cloud. Each token $\mathbf{t}_q$ is associated with the 2D coordinates $\mathbf{c}_q$ of the center of its corresponding pillar~$q$.

% ===============================================
\subsection{Vanilla ViT} 
\label{sec:method_vit}

The tokens $\mathbf{t}_q^{(0)}$, $q = 1, \ldots, Q$, are processed with a Vanilla ViT \cite{dosovitskiy2021image}, \alexRoundOne{for $\ell = 0, \ldots, L_{\rm vit} - 1$}:
\begin{align}
\tilde{\mathbf{T}}^{(\ell + 1)} 
    & = {\rm MSA}( {\rm LN}( \mathbf{T}^{(\ell)} ), \; \mathbf{C}) + \mathbf{T}^{(\ell)}
\\
\mathbf{T}^{(\ell + 1)} 
    & = {\rm MLP}( {\rm LN}( \tilde{\mathbf{T}}^{(\ell + 1)} ) ) + \tilde{\mathbf{T}}^{(\ell + 1)}
\end{align}
where $\mathbf{T}^{(0)} = [\mathbf{t}_1^{(0)}, \ldots, \mathbf{t}_Q^{(0)}]$, % and
$\mathbf{C} = [\mathbf{c}_1, \ldots, \mathbf{c}_Q]$ \renaud{are RoPE coordinates \cite{rope}}, and $L_{\rm vit}$ is the number of transformer blocks. 
Layer norm (LN) is applied before the multi-head self-attention (MSA) and the MLP. 
\renaudRoundOne{The} % For 
positional encoding in the MSA \renaudRoundOne{uses} %, we use 
a classical 2D RoPE, % by 
repeating % the 
1D RoPE % operation 
on each axis \cite{rope,heo2024ropevit}.
The MLP contains two layers with a GELU non-linearity.

% To allow 
\renaudRoundOne{For} batch processing of multiple point clouds, the token sequence length \renaudRoundOne{of} % for 
each point cloud is equalized to the maximum length in a batch (maximum number of non-empty pillars for a point cloud) by repeating a padding token. % the appropriate number of times.

% ===============================================
\subsection{Segmentation head} 
\label{sec:method_head}

% To allow 
\renaudRoundOne{For} semantic segmentation at % \renaudRoundOne{point level}, % 
the level of the points, 
the tokens $\mathbf{T}^{( L_{\rm vit})} = [\mathbf{t}_1^{( L_{\rm vit})}, \ldots, \mathbf{t}_Q^{( L_{\rm vit})}]$ are lifted from BEV to 3D: each point in a given pillar $q$ receives the corresponding token $\mathbf{t}_q^{( L_{\rm vit})}$. % This operation alone would be insufficient to 
\renaudRoundOne{To} differentiate the individual classes of % the 
points falling into the same pillar, each % . Each 
lifted feature vector is % thus 
augmented with the \renaudRoundOne{associated} % corresponding 
point embedding $\mathbf{p}^{(L_{\rm emb})}_i$.

With \renaudRoundOne{an} % a slight 
abuse of notation, let $\mathbf{t}_i^{( L_{\rm vit})}$ be the features lifted from BEV for point $i$. The final 
features $\mathbf{r}_i$ of % for point 
$i$ satisfies
\begin{align}
\tilde{\mathbf{r}}_i 
    & = 
    [g_{\rm vit}(\mathbf{t}^{(L_{\rm vit})}_i) \; , \; g_{\rm emb}(\mathbf{p}^{(L_{\rm emb})}_i)],
\\
\mathbf{r}_i 
    & = 
    g_{\rm final}\left( g_{\rm gate}(\tilde{\mathbf{r}}_i) \odot \tilde{\mathbf{r}}_i \right),
\end{align}
where $g_{\rm vit}$, $g_{\rm emb}$, $g_{\rm final}$ contain a linear layer followed by batch norm and ReLU, and have an output dimension set to the inner dimension $F_\ell$ of the tokenizer. 
\renaudRoundOne{To enable focusing more on the initial point embedding or on the representation coming out of the ViT, depending on the type of object to classify, we}
use a gating function $g_{\rm gate}$, made of a linear layer followed by a sigmoid, to reweigh the contribution of each channel of the two types of representations. 

The final point feature $\mathbf{r}_i$ then enters a linear % classification
layer in charge of classifying % the point 
into one of the semantic classes.

% % ===============================================
\subsection{Point cloud augmentation by mixing} 
\label{sec:method_aug}

Transformers in general, and ViTs in particular, may need large datasets to train well \cite{dosovitskiy2021image}. Data augmentation is thus a \renaudRoundOne{key} % major 
strategy \renaudRoundOne{to perform well} % for improving performance, 
especially in the automotive lidar setting where the number of point clouds is % often 
considerably smaller than the amount of images used to train ViTs.

\alexRoundOne{
Classical point cloud augmentations include random rotation, flipping and scaling.\,To further improve performance\rlap, PillarMix \cite{pillarmix} was introduced for semi-supervised object detection. It takes two point clouds, applies the above augmentations to each, divides them into regular square pillars, and combines them in a checkerboard arrangement. 
}

\alexRoundOne{
We propose a variant of PillarMix that yields better results for training our models (see \cref{sec:ablation}). 
Given a reference point cloud, we randomly select $N-1$ others and apply the usual rotate-flip-scale augmentations to all $N$. 
We then build a new point cloud pillar by pillar: for each occupied pillar in the reference, we check whether the same location is occupied in at least one of the other point clouds. If so, we randomly pick one among the up-to-$N$ occupied pillars at that location; otherwise, we keep the reference pillar.
}

The final point cloud is a mix of points from $N$ point clouds. % The 
\renaudRoundOne{Point cloud} truncation % of point clouds 
at pillar edges creates hard samples to learn from\renaudRoundOne{, and} % .  The 
mixing creates context variations around objects to segment. Note that, as spliced pillars are from the same location with respect to the ego vehicle, they are consistent regarding the scan pattern (viewpoint, density). They are also consistent % There is also a consistency 
at a % more
global level as the augmentation replaces an occupied pillar by another \renaudRoundOne{one,} % occupied pillar, 
thus \renaudRoundOne{retaining} % and thus retains 
the general shape of the reference point cloud.

The choice of the number $N$ of point clouds to mix and the \renaudRoundOne{pillar} size $S$ % of pillars 
is studied in \cref{sec:ablation}. Note that the size $S$ of \renaudRoundOne{mixed} pillars is much larger \renaud{(7.5\,m in our experiments)} % (several meters) % long) 
than the size $s$ of the pillars used for tokenization \renaud{(0.5\,m)}. % (less than one meter).

\renaudRoundOne{Our variant} \pmixplus differs from PillarMix in the following ways. With its checkerboard pattern, PillarMix only mixes two point clouds, while our ablations show (\cref{tab:aug_hyperparameters}) that mixing more than two is beneficial. Besides, when mixing point clouds \renaudRoundOne{from} % representing 
different road shapes, thus with entire areas occupied \renaudRoundOne{in only one point cloud,} % only in one of the two point clouds, 
the PillarMix checkerboard pattern creates pillars mostly surrounded by empty space (\renaudRoundOne{the} empty pillars from the other point cloud), similar to stand-alone crops. \renaudRoundOne{It reduces} % , thus reducing 
the variety of \renaudRoundOne{data found in the immediate pillar environment}. % information coming the immediate environment of the pillar. 
In contrast, as \pmixplus\  preserves pillar occupancy, it can make the most of context variation. Last, the number of occupied pillars in PillarMix can be larger than in the reference point cloud, \renaudRoundOne{thus} % therefore 
slightly increasing the number of tokens and the computational load during training. 
With \pmixplus, the number of occupied pillars stays the same, % hence 
almost preserving the number of tokens \renaudRoundOne{(up to alignment with tokenization grid)}.

% ===============================================
%
% ===============================================
\section{Experiments}
\label{sec:}

In this section, we evaluate the 
%% semantic segmentation 
performance of % our proposed 
VaViT backbones for sizes S, B and L, across three lidar datasets.

\begin{table*}[p]
\small
\ra{1.3}
\newcommand*\rotext{\multicolumn{1}{R{65}{1em}}}
\setlength{\tabcolsep}{3pt}
\renewcommand{\arraystretch}{1.2}
\centering
\begin{subtable}[b]{\linewidth}
\centering
\begin{tabular}{l@{}r c | c c c c c c c c c c c c c c c c}
\toprule 
\raisebox{8ex}{\rlap{\textbf{nuScenes val set}}}%
\raisebox{5.5ex}{\rlap{\textbf{no TTA}}}%
Method
    &
    & \rotext{mIoU\%}
    & \rotext{barrier}
    & \rotext{bicycle}
    & \rotext{bus}
    & \rotext{car}
    & \rotext{const. veh.}
    & \rotext{motorcycle}
    & \rotext{pedestrian}
    & \rotext{traffic cone}
    & \rotext{trailer}
    & \rotext{truck}
    & \rotext{driv. surf.}
    & \rotext{other flat}
    & \rotext{sidewalk}
    & \rotext{terrain}
    & \rotext{manmade}
    & \rotext{vegetation}
\\
\midrule
GFNet 
    & \cite{gfnet}
    & 76.1
    & \textbf{81.1}
    & 31.6 
    & 76.0 
    & 90.5 
    & 60.2
    & 80.7 
    & 75.3 
    & \uline{71.8}
    & \textbf{82.5}
    & 65.1 
    & \textbf{97.8}
    & 67.0 
    & \textbf{80.4}
    & 76.2
    & \textbf{91.8}
    & 88.9
\\
Cylinder3D 
    & \cite{Cylinder3d}
    & 76.1
    & 76.4 
    & 40.3 
    & 91.2 
    & 93.8
    & 51.3
    & 78.0 
    & 78.9 
    & 64.9 
    & 62.1 
    & 84.4
    & 96.8 
    & 71.6
    & 76.4
    & 75.4
    & 90.5 
    & 87.4
\\
RPVNet 
    & \cite{RPVNet}
    & 77.6
    & 78.2
    & 43.4
    & 92.7
    & 93.2
    & 49.0
    & 85.7
    & 80.5
    & 66.0
    & 66.9
    & 84.0
    & 96.9
    & 73.5
    & 75.9
    & 76.0
    & 90.6
    & 88.9
\\
WI-384 
    & \cite{waffleiron}
    & 77.6
    & 78.7
    & 51.3
    & 93.6 
    & 88.2
    & 47.2
    & 86.5
    & 81.7
    & 68.9
    & 69.3
    & 83.1
    & 96.9
    & 74.3
    & 75.6 
    & 74.2
    & 87.2 
    & 85.2
\\
SDSeg3D 
    & \cite{sdseg3d}
    & 77.7
    & 77.5 
    & 49.4 
    & 93.9 
    & 92.5 
    & 54.9
    & 86.7 
    & 80.1 
    & 67.8 
    & 65.7 
    & 86.0
    & 96.4 
    & 74.0
    & 74.9
    & 74.5
    & 86.0
    & 82.8
\\
\rowcolor{green!15}
\renaud{VaUNet-B}
    & \llap{\renaud{(ours)}}
    & 77.9
    & 78.6
    & 30.9
    & 95.3
    & 93.1
    & 51.1
    & 85.0
    & 80.9
    & 68.8
    & 71.7 
    & 84.2 
    & 96.9 
    & \textbf{77.0} 
    & 76.5 
    & \uline{76.7} 
    & 91.0 
    & 89.5
\\
RangeFormer 
    & \cite{rangeformer}
    & 78.1 
    & 78.0 
    & 45.2 
    & 94.0 
    & 92.9 
    & 58.7 
    & 83.9 
    & 77.9 
    & 69.1 
    & 63.7 
    & 85.6 
    & 96.7 
    & 74.5 
    & 75.1 
    & 75.3 
    & 89.1 
    & 87.5
\\
SphereFormer
    & \cite{sphereformer}
    & 78.4 
    & 77.7 
    & 43.8 
    & 94.5 
    & 93.1 
    & 52.4 
    & 86.9 
    & 81.2 
    & 65.4 
    & 73.4 
    & 85.3 
    & 97.0 
    & 73.4 
    & 75.4 
    & 75.0 
    & 91.0 
    & 89.2
\\
FlatFormer-S$^\dagger$ 
    & \cite{flatformer}
    & 78.6
    & 78.8
    & 45.4
    & 94.5
    & 93.0
    & 58.1
    & 81.5
    & 79.9
    & 67.2
    & 71.5
    & 85.2
    & 96.8
    & 74.5
    & 75.9
    & 75.4
    & 90.8
    & 89.2
\\
WI-768 
    & \cite{scalr}
    & 78.7
    & 79.2 
    & \uline{53.2} 
    & 92.5
    & 88.1
    & 50.4
    & \uline{87.8}
    & \uline{83.4}    
    & 70.3     
    & 73.5    
    & 84.1
    & 96.9
    & 73.3
    & 75.5
    & 75.4
    & 89.9
    & 86.5
\\
PTv3$^\dagger$
    & \cite{ptv3}
    & 78.9
    & \uline{79.7}
    & 49.7
    & \uline{95.6}
    & 93.6
    & 46.0
    & 84.6
    & 82.2
    & 70.1
    & 72.9
    & 84.0
    & 96.9
    & 74.7
    & 76.3
    & 75.5
    & 90.8
    & 89.1
\\
LitePT$^\dagger$
    & \cite{litept}
    & \underline{80.7}
    & 79.2
    & 52.3
    & \textbf{95.7}
    & \textbf{94.2}
    & \uline{64.4}
    & 84.8
    & 82.8
    & 68.9
    & 72.9
    & \textbf{87.7}
    & 97.0
    & \uline{76.7}
    & \uline{77.3}
    & \textbf{76.8}
    & 91.1
    & \uline{89.6}
\\
\rowcolor{blue!15}
VaViT-B 
    & \llap{(ours)}
    & \bf 81.3
    & 79.6
    & \textbf{54.4}
    & 95.2
    & \uline{93.9}
    & \textbf{65.4}
    & \textbf{88.8}
    & \textbf{83.6}
    & \textbf{72.0}
    & \uline{74.1}
    & \uline{87.3}
    & \uline{97.1}
    & 75.3
    & 77.0
    & 76.2
    & \uline{91.3}
    & \textbf{89.7}
\\
\bottomrule
\end{tabular}
\caption{Performance on nuScenes \emph{validation} set, \emph{without TTA}.}
\label{tab:nuscenes_val_set}
\end{subtable}

\medskip
%%%%%%%%%%%%%%%%%%%%%%%%%%%%%%%%%%
\begin{subtable}[b]{\linewidth}
\centering
\begin{tabular}{l c | c c c c c c c c c c c c c c c c c c c}
\toprule 
\raisebox{9.5ex}{\rlap{\textbf{SemanticKITTI val set}}}%
\raisebox{7ex}{\rlap{\textbf{no TTA}}}%
Method
    & \rotext{mIoU\%}
    & \rotext{car}
    & \rotext{bicycle}
    & \rotext{motorcycle}
    & \rotext{truck}
    & \rotext{other-vehicle}
    & \rotext{person}
    & \rotext{bicyclist}
    & \rotext{motorcyclist}
    & \rotext{road}
    & \rotext{parking}
    & \rotext{sidewalk}
    & \rotext{other-ground}
    & \rotext{building}
    & \rotext{fence}
    & \rotext{vegetation}
    & \rotext{trunk}
    & \rotext{terrain}
    & \rotext{pole}
    & \rotext{traffic-sign}
\\
\midrule
MinkUNet$^\diamond$
    & 63.8
    &-&-&-&-&-&-&-&-&-&-&-&-&-&-&-&-&-&-&-
\\
Cylinder3D$^\diamond$
    & 64.3
    &-&-&-&-&-&-&-&-&-&-&-&-&-&-&-&-&-&-&-
\\
SPVNAS$^\diamond$
    & 64.7
    &-&-&-&-&-&-&-&-&-&-&-&-&-&-&-&-&-&-&-
\\
FlatFormer-S$^\dagger$ % lr 0.0002
    & 65.3
    & 94.9 
    & \uline{56.8}
    & 75.8
    & 88.0
    & 40.7
    & 77.3
    & 91.6
    & 0.0
    & 94.5
    & 44.9
    & 82.7
    & 0.2
    & 90.6
    & 60.7
    & 87.2
    & 69.5
    & 72.0
    & 64.8
    & 48.5
\\
PTv3$^\dagger$ 
    & 66.2
    & \textbf{96.2}
    & 45.2
    & \uline{79.0}
    & 81.9
    & 55.6
    & 79.0
    & 91.0
    & 0.0
    & 95.3
    & 53.2
    & \textbf{84.4}
    & 0.3
    & 90.3
    & 60.4
    & 87.6
    & 70.3
    & 73.0
    & \textbf{66.8}
    & 48.2
\\
\rowcolor{green!15}
\renaud{VaUNet-B}
    & 67.0
    & 95.6 & 52.0 & 69.4 & 94.6 & 53.5 & 78.3 & 90.6 & 0.0 & 95.5 & \textbf{61.4} & \uline{83.9} & 0.5 & 90.5 & 60.7 & \textbf{88.7} & 69.2 & \textbf{75.3} & 63.1 & \uline{49.9}
\\
SphereFormer$^\diamond$
    & 67.8
    &-&-&-&-&-&-&-&-&-&-&-&-&-&-&-&-&-&-&-
\\
WI-256
    & 68.0
    & \uline{96.1} 
    & \textbf{58.1}
    & \textbf{79.7}
    & 77.4
    & \textbf{59.0} 
    & 81.1
    & \uline{92.2} 
    & \textbf{1.3}
    & 95.5 
    & 50.2 
    & 83.6 
    & \textbf{6.0} 
    & \textbf{92.1} 
    & \textbf{67.5}
    & 87.8 
    & \textbf{73.8} 
    & 73.0
    & \uline{65.7} 
    & \textbf{52.2}
\\

\rowcolor{blue!15}VaViT-B
    & \uline{68.1}
    & 95.7 
    & 55.1 
    & \uline{79.0} 
    & \textbf{95.6} 
    & \uline{56.8} 
    & \uline{81.6} 
    & \textbf{94.6} 
    & \uline{0.0} 
    & \textbf{95.9} 
    & 52.3 
    & \uline{83.9} 
    & 0.7 
    & \uline{91.1} 
    & \uline{64.0} 
    & 88.2 
    & 72.9 
    & 73.8 
    & \uline{65.7} 
    & 48.9
\\

\rowcolor{blue!15}VaViT-B*
    & \textbf{68.4}
    & 95.7 
    & 54.4 
    & 78.4 
    & \uline{94.9} 
    & \uline{56.8} 
    & \textbf{81.8} 
    & \textbf{94.6} 
    & \uline{0.0} 
    & \uline{95.7} 
    & \uline{54.5} 
    & 83.8 
    & \uline{3.9} 
    & 90.9 
    & 63.8 
    & \uline{88.4} 
    & \uline{73.1} 
    & \uline{74.2} 
    & 65.1 
    & 49.1
\\

\bottomrule
\end{tabular}
\caption{Performance on SemanticKITTI \emph{validation} set, \emph{without TTA}. $^\diamond$from \cite{ptv3}. \renaud{LitePT does not evaluate on SemanticKITTI.}}
\label{tab:semkitti_val_set}
\end{subtable}

\medskip
%%%%%%%%%%%%%%%%%%%%%%%%%%%%%%%%%%%
\begin{subtable}[b]{\linewidth}
\centering
\setlength{\tabcolsep}{1.7pt}
\begin{tabular}{l c | c c c c c c c c c c c c c c c c c c c c c c}
\toprule 
\raisebox{9ex}{\rlap{\textbf{WOD val set}}}%
\raisebox{6.5ex}{\rlap{\textbf{no TTA}}}%Method
    & \rotext{mIoU\%}
    & \rotext{car}
    & \rotext{truck}
    & \rotext{bus}
    & \rotext{oth.~veh.}
    & \rotext{motorcyclist}
    & \rotext{bicyclist}
    & \rotext{pedestrian}
    & \rotext{sign}
    & \rotext{traffic light}
    & \rotext{pole}
    & \rotext{const.~cone}
    & \rotext{bicycle}
    & \rotext{motorcycle}
    & \rotext{building}
    & \rotext{vegetation}
    & \rotext{trunk}
    & \rotext{curb}
    & \rotext{road}
    & \rotext{lane marker}
    & \rotext{other ground}
    & \rotext{walkable}
    & \rotext{sidewalk}
\\
\midrule
SphereFormer
    & 69.9
    & 94.5 
    & 61.6 
    & \textbf{87.7}
    & 40.2 
    & \phantom{0}0.9 
    & 69.7 
    & 90.2 
    & 73.9 
    & \textbf{41.8} 
    & 77.2 
    & 65.4 
    & 71.9 
    & 83.7 
    & \uline{95.9} 
    & 91.7 
    & 68.4 
    & 69.8 
    & 93.3 
    & 53.9 
    & 47.9 
    & 80.8 
    & 77.2
\\
PTv3$^\dagger$
    & 70.2
    & \textbf{94.8}
    & \textbf{63.3}
    & 83.9
    & 34.6
    & \phantom{0}3.9
    & \uline{77.1}
    & 90.4
    & 70.8
    & 32.7
    & 77.3
    & 67.1
    & \uline{74.3}
    & \textbf{88.0}
    & 95.8
    & 91.7
    & 67.5
    & 70.8
    & \uline{93.4}
    & \textbf{56.6}
    & \textbf{52.0}
    & \textbf{81.5}
    & 77.8
\\
\rowcolor{green!15}
\renaud{VaUNet-B}
    & 70.3
    & 94.5 & 60.9 & 86.3 & \textbf{47.9} & \phantom{0}5.1 & 76.5 & 89.7 & 73.8 & 34.2 & 78.0 & \uline{69.8} & 74.0 & 73.3 & 95.8 & 91.4 & 68.7 & 69.8 & 93.2 & 55.1 & 50.2 & 80.6 & \uline{77.9}
\\
\rowcolor{blue!15}VaViT-B 
    & 70.5
    & 94.4 
    & 61.3
    & 84.2
    & \uline{42.8}
    & \textbf{10.2}
    & 74.7 
    & \textbf{90.8} 
    & \uline{74.4}
    & 33.7 
    & 79.5 
    & 63.3 
    & 73.8 
    & 80.6
    & \textbf{96.1} 
    & \textbf{92.0} 
    & 69.9 
    & \textbf{71.1} 
    & \textbf{93.5} 
    & 55.4 
    & 50.3
    & \uline{81.4} 
    & \textbf{78.3}
\\
\rowcolor{blue!15}VaViT-B*
    & \underline{70.9}
    & 94.5 
    & 62.5 
    & \uline{86.7} 
    & 39.3 
    & \phantom{0}7.7 
    & 76.9 
    & \textbf{90.8} 
    & 74.0 
    & 33.6 
    & \textbf{79.6} 
    & 66.6 
    & \textbf{74.4} 
    & 87.2 
    & 96.0 
    & 91.9 
    & \textbf{70.2} 
    & 70.8 
    & 93.4 
    & 55.1 
    & 49.9 
    & \uline{81.4} 
    & \textbf{78.3}
\\
LitePT$^\dagger$
    & \bf 71.7
    & \uline{94.6}
    & \uline{63.1}
    & 83.9
    & 41.2
    & \phantom{0}\uline{9.8}
    & \textbf{78.4}
    & \uline{90.7}
    & \textbf{75.0}
    & \uline{39.5}
    & \uline{79.0}
    & \textbf{74.5}
    & 73.6
    & \uline{87.6}
    & \uline{95.9}
    & \uline{91.8}
    & \uline{70.0}
    & \uline{69.8}
    & 93.2
    & \uline{56.3}
    & \uline{51.1}
    & 81.0
    & 77.3
\\
\bottomrule
\end{tabular}
\caption{Performance on the WOD \emph{validation} split, \emph{without TTA}.}
\label{tab:waymo_val_set}
\end{subtable}
\vspace*{-3.5ex}
\caption{\textbf{Performance on nuScenes, SemanticKITTI and WOD validation sets, without TTA.} \renaud{Models are trained from scratch on each individual dataset.}
$^\dagger$%\,indicates 
results that we reproduced from the original code. 
$^\diamond$results from \cite{ptv3}.
$^*$results of best epoch rather than last epoch (this information is actually unknown for many publications).
Best and second-best mIoUs are indicated respectively by \textbf{bold} and \uline{underlining}. 
}
\label{tab:all_val_set}
\end{table*}

% ===============================================
\subsection{Experimental setup}
\label{sec:exp_setup}

\paragraph{Datasets.} 
%\subsubsection{Datasets.} 
% 
We evaluate our method on three benchmark datasets: nuScenes \cite{nuscenes}, SemanticKITTI \cite{semkitti}, and the Waymo Open Dataset (WOD) \cite{waymo}. The nuScenes dataset \renaudRoundOne{was} % consists of 1,000 scenes 
captured in Boston and Singapore, with point clouds annotated into 16 semantic classes. SemanticKITTI (SemKITTI) \renaudRoundOne{was} % comprises 22 \renaudRoundOne{longer} sequences 
recorded in Karlsruhe, Germany, with 19 semantic classes. Finally, WOD includes scenes from San Francisco, Phoenix and Mountain View, with semantic masks for 22 % distinct 
classes.

\paragraph{VaViT backbones.}
We experiment with three versions of our backbone: VaViT-S, VaViT-B, VaViT-L, leveraging the ViT-S, ViT-B or ViT-L architectures \cite{dosovitskiy2021image,dino}, respectively. The feature size $F_\ell$ is set to 64\,/128\,/\,192 for the S\,/\,B\,/\,L versions. This size was set by conducting a sensitivity study on VaViT-S (see \cref{sec:ablation}) and the size was scaled proportionally to the inner dimension of the ViTs. Unless specified otherwise, we use $L_{\rm emb} = 4$ layers for tokenization.

\paragraph{Baselines.}
The closest baseline \renaudRoundOne{method}s are PTv3~\cite{ptv3}, LitePT~\cite{litept}, and FlatFormer~\cite{flatformer}. For PTv3, we were able to reproduce the original results on nuScenes and WOD. On SemanticKITTI, reproducing the original \renaudRoundOne{result} is known to be difficult\footnote{https://github.com/Pointcept/Pointcept/issues/186} \cite{dinointheroom}. We were able to reach an mIoU of $68.8\%$ with test time augmentation (TTA) by adapting the configuration provided for nuScenes. This score is in line with the reproduced result in \cite{dinointheroom}. For LitePT, 
we used the publicly released checkpoints.
\renaud{which do not cover SemanticKITTI.} 
FlatFormer is originally designed for object detection. We repurposed it for semantic segmentation as follows: we extracted the transformer architecture working in BEV and inserted it between our tokenizer and our semantic segmentation head. We are thus not reusing the tokenizer of FlatFormer, which was in any case designed and tuned for object detection in BEV, not for semantic segmentation. The resulting adaptation of FlatFormer for semantic segmentation is denoted FlatFormer-S (details in appendix). \renaud{VaUNet replaces the ViT in VaVit by a U-Net (\cref{sec:ablations}).}

\paragraph{Test-time augmentation (TTA).} 
The latest works in point cloud semantic segmentation report scores %% \renaudRoundOne{after} % by 
resorting to \renaudRoundOne{TTAs}. % test-time augmentations. 
\renaudRoundOne{This} % It is a 
practice % that 
has evolved over time. We argue that \renaudRoundOne{it} % this setting 
does not fully \renaudRoundOne{}{demonstrate} % show
the properties of the backbones. Besides, it is impractical for a number of real-world uses. \renaudRoundOne{In} % As a matter of 
fact, a full evaluation with TTA on the considered validation sets takes several hours for PTv3 and LitePT when using a single GPU. In contrast, it is \renaudRoundOne{only} a matter of \renaudRoundOne{a few} minutes for our VaViT models without TTA.
In our main tables, we therefore report results \emph{without} TTA (reproducing results when necessary) and discuss TTA in \cref{sec:tta}.

\subsection{Comparison to the state of the art}
\label{sec:exp_main}

\paragraph{\renaud{Segmentation quality.}}

\renaud{We report in \cref{tab:all_val_set} the semantic segmentation performance of various models trained from scratch individually on the respective training set of nuScenes \cite{nuscenes}, SemanticKITTI \cite{semkitti} and WOD \cite{waymo}, and evaluated on their respective validation set, without TTA}. While training stochasticity generally has little impact on the final performance with nuScenes, training on SemanticKITTI, or WOD, may display some % visible 
variance. As the practices differ in the literature regarding citing the performance at \renaudRoundOne{the} last or best epoch, we report for SemanticKITTI and WOD not only VaViT results for our final training epoch but also for the best epoch, denoted VaViT$^*$. \renaud{A variance study with 3 runs on SemanticKITTI is in appendix.}

\begin{table}[t]
\small
\centering
\setlength{\tabcolsep}{3pt}
\resizebox{0.99\linewidth}{!}{\renaud{\begin{tabular}{l|ccc}
\toprule
         & nuScenes                 & SemanticKITTI             & WOD             \\ 
\midrule
PTv3     & 59\,ms (50\,ms) / 0.6 GB & 111\,ms (87\,ms) / \textbf{1.5 GB} & 142\,ms (82\,ms) / 1.0 GB \\
LitePT-S & 46\,ms (35\,ms) / \textbf{0.5 GB} & \textit{no code}          & \textbf{104\,ms} (48\,ms) / \textbf{0.9 GB} \\
VaViT-B  & \textbf{17\,ms} (17\,ms) / 0.7 GB &  \textbf{70\,ms} (70\,ms) / 1.7 GB & 109\,ms (109\,ms) / 1.9 GB \\
\bottomrule
\end{tabular}}}
\vspace{-7pt}
\caption{\renaud{\textbf{Evaluation of semantic segmentation efficacy} for a single forward pass (no TTA), with batch size 1, on a Nvidia H100 GPU. Latency measurements exclude I/Os, dataloader preprocessing, and are made after warm-up. The number in parentheses is the time spent only in the GPU, thus excluding I/Os and CPU.}}
\label{tab:runningtimes}
\end{table}

On nuScenes, VaViT-B achieves state-of-the-art performance surpassing the recent LitePT. Interestingly, VaViT-B ranks first on traffic cone and bicycle, despite their small sizes and the coarse grid used for tokenization, \renaud{showing the expressiveness of our 3D embeddings despite their BEV projection on a coarse 2D grid.} \renaud{On SemanticKITTI (for which there is no LitePT code nor evaluation), and to a lesser extent on WOD, our models also compete with today's best methods, \renaud{including again on small objects like} 
bicycle, person/pedestrian, motorcycle, or traffic sign.
VaViT is good too (often SOTA or second best) with large objects (e.g., road, building), possibly thanks to global attention.}

\paragraph{\renaud{Latency and memory consumption.}}

\renaud{\cref{tab:runningtimes} reports the running time and memory usage of a single forward pass on a point cloud. VaViT is about 3$\times$ faster than PTv3 and LitePT on nuScenes. Because of its powerful point embedding, VaViT is a bit more affected than the other methods by the denser lidars of SemanticKITTI (about 2$\times$ denser) and WOD (about 4$\times$ denser), but it remains faster than PTv3, though not with such a large margin. On WOD, VaViT and LitePT have comparable latency, better than PTv3's.}

\renaudRoundOne{Note however that the published mIoUs for most methods actually rely on multiple passes, using TTAs and sometimes so-called ``chunking'', whose details often cannot be found in the publications. For chunking, after a first decimation (typically with 2.5\,cm voxels), points in actual inference voxels (typically with 5\,cm-voxels) are sampled (typically up to 8 times) to create several point cloud variants to process and ensemble. And this is typically performed for each TTA, hence multiplying further the number of passes \renaud{(e.g., $\text{12}\,{\times}\,\text{8}\,{=}\,\text{96}$ for PTv3 on nuScenes)}. This significantly increases the actual latency and memory usage at test time \renaud{(see appendix)}. While VaViT could use TTAs and chunking, we do not consider it an integral part of the method as it has empirically little impact of the performance (\cref{sec:tta}).}

\renaud{
The three methods have a comparable runtime footprint on nuScenes. With denser lidars, all methods require 2$\times$ to 3$\times$ more memory. On WOD, VaViT requires 1.9\,GB, which remains moderate, while LitePT scales better to 0.9\,GB.}

% ===============================================
\subsection{Ablation and sensitivity study}
\label{sec:ablations}

In this section, we conduct ablation experiments to validate the design of our tokenizer and segmentation head, \renaud{evaluate the benefit of using a plain ViT vs a U-Net,} and investigate the sensitivity of VaViT to various hyperparameters, data augmentations and capacity scaling. 

All experiments in this section are performed using VaViT-S unless otherwise mentioned. The default setting is $L_{\rm emb}\,=\,4$, $F_\ell\,=\,64$, token pillar resolution $s = 1\,\text{m}$ for tokenization, $N\,=\,3$ and $S\,=\;5\,\text{m}$ 
in \pmixplus. 
\label{sec:ablation}

\begin{table}[t]
\centering
\small
\ra{1.2}
\setlength{\tabcolsep}{6pt}
\renewcommand{\arraystretch}{1.1}
\begin{tabular}{cccccc}
\toprule
Mean 
    &  Max
    & Token gat.
    & Seg.\,gat.
    & \multicolumn{2}{c}{mIoU\%}
    % & 
\\[-0.5mm]
pool 
    & pool
    & $f_{\rm gate}(\cdot)$
    & $g_{\rm gate}(\cdot)$
    & \!SemKITTI\!
    & WOD
\\
\cmidrule(lr){1-4}
\cmidrule(lr){5-6}
\cmark 
    & -
    & \xmark
    & \xmark
    & 63.5
    & 66.9
\\
-
    & \cmark 
    & \xmark
    & \xmark
    & 63.6
    & 66.9
\\
-
    & \cmark 
    & \cmark
    & \xmark
    & 65.3
    & 68.9
\\
\rowcolor{blue!15}-
    & \cmark 
    & \cmark
    & \cmark
    & \bf 66.0
    & \bf 69.2
\\
\bottomrule
\end{tabular}    
\vspace{-7pt}
\caption{\textbf{Ablations for the tokenizer and segmentation head.} We compare BEV projection pooling strategies and evaluate our gating mechanisms on the val sets of SemKITTI and WOD.
}
\label{tab:architecture}
\vspace{-7pt}
\end{table}
\begin{table}[t]
\centering
\begin{minipage}[t]{0.31\linewidth}
\centering
\scriptsize
\ra{1.1}
\setlength{\tabcolsep}{3pt}
\begin{tabular}[t]{ccc}
\toprule
$F_\ell$ 
    & SemKITTI
    & WOD
\\
\cmidrule(lr){1-1}
\cmidrule(lr){2-3}
\phantom{1}16  
    & 63.6  
    & 66.3 
\\
\phantom{1}32	
    & 64.8  
    & 67.9 
\\
\rowcolor{blue!15}\phantom{1}64	
    & 66.0  
    & \bf 69.2 
\\
\phantom{1}96	
    & 66.0  
    & 68.5 
\\
128	
    & \bf 66.1  
    & 68.9 
\\
\bottomrule
\end{tabular}
\vspace{1mm}
\\
(a) $L_{\rm emb}=4$; $s=1$\,m. 
\end{minipage}
\hspace*{-7.5pt}\hfill
\begin{minipage}[t]{0.31\linewidth}
\centering
\scriptsize
\ra{1.1}
\setlength{\tabcolsep}{3pt}
\begin{tabular}[t]{ccc}
\toprule
$L_{\rm emb}$ 
    & SemKITTI
    & WOD
\\
\cmidrule(lr){1-1}
\cmidrule(lr){2-3}
1	
    & 64.2	
    & 67.2    
\\
2	
    & 64.4
    & 68.3    
\\
3	
    & 65.6	
    & 68.3    
\\
\rowcolor{blue!15} 4	
    &	\bf 66.0	
    &   \bf 69.2
\\
5	
    & 65.2	
    & 68.3
\\
\bottomrule
\end{tabular}
\vspace{1mm}
\\
(b) $F_\ell = 64$; $s=1$\,m. 
\end{minipage}
\hfill
\begin{minipage}[t]{0.31\linewidth}
\centering
\scriptsize
\ra{1.1}
\setlength{\tabcolsep}{3pt}
\begin{tabular}[t]{ccc}
\toprule
$s$ (m) 
    & SemKITTI
    & WOD
\\
\cmidrule(lr){1-1}
\cmidrule(lr){2-3}
\rowcolor{blue!15} 0.50	
    & \bf 67.1 
    & \bf 69.4
\\
0.75    
    & 66.6 
    & 69.3
\\
1.00    
    & 66.0 
    & 69.2
\\
1.25
    & 64.8 
    & 68.4
\\
\bottomrule
\end{tabular}
\vspace{1mm}
\\
(c) $L_{\rm emb}=4$; $F_\ell = 64$.
\end{minipage}
\vspace{-5pt}
\caption{\textbf{Sensitivity analysis of architecture hyperparameters.} We study the impact of (a)~the feature dimension $F_\ell$ , (b)~the number $L_{\rm emb}$ of layers
% in the tokenizer, 
and (c)~the pillar resolution $s$ for the tokenizer. 
}
\label{tab:arch_hyperparameters}
\vspace{-5pt}
\end{table}

\paragraph{Tokenizer and segmentation head architectures.} 
We justify our architectural choices in \cref{tab:architecture}. Starting from a baseline that utilizes mean pooling for BEV projection as used in \cite{waffleiron}, we find that switching to max pooling yields comparable performance; we adopted the latter in the final models. Furthermore, the results demonstrate that incorporating our proposed gating mechanisms into both the tokenizer and the segmentation head provide a consistent performance boost, confirming their relevance. % in refining spatial features.

\paragraph{Architecture hyperparameters.} 
We explore the impact of $F_\ell$, $L_{\rm emb}$ and $s$ in \cref{tab:arch_hyperparameters}. \cref{tab:arch_hyperparameters}a shows that the optimal width
is $F_\ell = 64$; performance degrades at lower values and plateaus beyond this point. \cref{tab:arch_hyperparameters}b shows that $L_{\rm emb} = 4$ yields the best results. Finally, \cref{tab:arch_hyperparameters}c indicates that segmentation performance improves as pillar resolution increases (smaller % pillar
size). We set our resolution limit at $s = \text{0.5}$\,m, as memory consumption begins to scale prohibitively below this threshold. \renaud{Besides, performance then mostly saturates: mIoU on nuScenes validation set without TTA only gains +0.1 pt when decreasing $s$ to 0.25\,m.}

\begin{table}[t]

\centering
\begin{minipage}[t]{0.31\linewidth}
\centering
\scriptsize
\ra{1.1}
\setlength{\tabcolsep}{2pt}
\begin{tabular}[t]{ccc}
\toprule
$S$ (m)
    & SemKITTI
    & WOD
\\
\cmidrule(lr){1-1}
\cmidrule(lr){2-3}
\clap{\renaud{no\,aug\rlap.}}
    & 62.3	
    & 64.9 
\\
\phantom{0}2.5	
    & 63.2	
    & 66.3	
\\
\phantom{0}5.0
    & 63.5
    & 66.9  
\\
\rowcolor{blue!15}\phantom{0}7.5
    & \bf 63.8	
    & \bf 67.1
\\
10.0
    & 63.2	
    & 66.5
\\
\bottomrule
\end{tabular}
\vspace{1mm}
\\
% (a) \pmixplus - $N = 3$
(a) Pillar size ($N = 3$)
\end{minipage}
\hspace{5pt}\hfill
\begin{minipage}[t]{0.35\linewidth}
\centering
\scriptsize
\ra{1.1}
\setlength{\tabcolsep}{2pt}
\begin{tabular}[t]{ccc}
\toprule
$N$
    & SemKITTI
    & WOD
\\
\cmidrule(lr){1-1}
\cmidrule(lr){2-3}
2   
    & 63.7	
    & 66.4 
\\
\rowcolor{blue!15}3   
    & \bf 63.8	
    & \bf 67.1 
\\
4   
    & 63.8	
    & 66.9
\\
5   
    & 63.4	
    & 67.1 
\\
\midrule
\clap{PillarMix}
    & 63.7
    & 66.3
\\
\bottomrule
\end{tabular}
\vspace{1mm}
\\
% (b) \pmixplus - $S = 7.5 \text{m}$
(b) No.\,scans ($S$\,=\,7.5\,m)
\end{minipage}
\hspace*{-10pt}
\hfill
\begin{minipage}[t]{0.31\linewidth}
\centering
\scriptsize
\ra{1.1}
\setlength{\tabcolsep}{2pt}
\begin{tabular}[t]{ccc}
\toprule
$N$
    & SemKITTI
    & WOD
\\
\cmidrule(lr){1-1}
\cmidrule(lr){2-3}
2   & 62.9	& 66.0 \\
3   & \bf 63.3	& \bf 66.0 \\
4   & 62.5	& 65.6 \\
5	& 62.6	& 66.1 \\
\bottomrule
\end{tabular}
\vspace{1mm}
\\
(c) PolarMix \cite{polarmix}
\end{minipage}
\vspace{-7pt}
\caption{\textbf{Hyperparameter study for \pmixplus and PolarMix.} We evaluate \pmixplus hyperparameters in (a) and (b), finding that mixing $N = 3$ point clouds with a pillar size $S$ of 7.5 m yields the best performance. In~(b) and (c), \pmixplus outperforms both PillarMix \cite{pillarmix} and PolarMix \cite{polarmix}, justifying our improvement.
}
\label{tab:aug_hyperparameters}
%\vspace{-2pt}
\end{table}

\paragraph{Hyperparameter choices \spyrosRoundOne{of \pmixplus}.}
We now justify our choice of hyperparameters for 
\pmixplus 
% \alexRoundOne{our PillarMix variant}
in \cref{tab:aug_hyperparameters}a-b. These experiments are done starting from the baseline setting in \cref{tab:architecture}, i.e., with mean pooling and without the gating mechanisms. We notice that the best results are obtained by mixing pillars of size $S = 7.5~\text{m}$ and considering $N\,=\,3$ point clouds for mixing. 

\paragraph{Comparison \spyrosRoundOne{of \pmixplus} to vanilla PillarMix and PolarMix.}
\alexRoundOne{
The last row in \cref{tab:aug_hyperparameters}b shows that the
improved version \pmixplus 
performs better than the original PillarMix in \cite{pillarmix}.
}
We also compare \pmixplus
against PolarMix \cite{polarmix} as a potential alternative in \cref{tab:aug_hyperparameters}c. In our implementation, PolarMix partitions the scene into $N$ random angular sectors for mixing. While mixing three point clouds also yields the best performance for PolarMix, its overall scores remain lower than those achieved with %\alexRoundOne{our variant.}
\pmixplus.

\paragraph{\renaud{Effect of the ViT backbone.}} \renaud{To show that our performance does not only come from our embedding, segmentation head and improved \pmixplus augmentation, we replaced the ViT backbone of VaViT by a dense 4-level 2D U-Net (lines ``VaUNet-B'' in \cref{tab:all_val_set}), adjusting the first level dimension to approximately match the number of parameters of VaViT-B (83M vs 86M). VaViT outperforms VaUNet on all three datasets (from 0.2 to 3.4 mIoU pt), showing that part of the gain stems from the use of a plain transformer.}

\begin{table}[t]

\small
\centering
\ra{1.1}
\setlength{\tabcolsep}{8pt}
\begin{tabular}{lccc}
\toprule 
Method
    & nuScenes
    & SemKITTI
    & WOD
\\
\midrule
VaViT-S
    & 80.6
    & \bf 67.6
    & 70.1
\\
\rowcolor{blue!15}VaViT-B
    & \bf 81.3	
    & \bf 67.6
    & \bf 70.5
\\
VaViT-L
     & 81.0
     & \bf 67.6
     & 70.2
\\
\bottomrule
\end{tabular}
\vspace{-7pt}
\caption{\textbf{Effect of model capacity.} mIoU saturates at VaViT-B.
}
\label{tab:capacity}
\end{table}

\paragraph{Capacity of VaViT models.}\label{sec:capacity}
We experimented with different model capacities: the VaViT-S, -B and -L models have respectively 21M, 86M and 304M parameters. Results in \cref{tab:capacity} shows that increasing the model capacity from VaViT-S to VaViT-B improves the performance, except on SemanticKITTI where its stagnates. This is probably partly explained by the fact that SemanticKITTI is the smallest dataset among all three. Increasing the capacity further up to VaViT-L does not improve performance. We hypothesize that the scale of these datasets is not large enough to witness an improved performance with VaViT-L \renaud{(which may overfit)}, or that even stronger augmentations are required. \renaud{Compared to variety in image datasets, automotive lidar datasets consist of recorded sequences, with little variation between successive frames. Moreover, some sequences show little diversity, e.g., on highways or in ``uniform'' neighborhoods.}

\renaud{To probe our hypothesis and experiment on more data, we associated the two datasets SemanticKITTI and WOD. To avoid issues with class differences and multi-dataset training, we actually finetuned SemanticKITTI starting from WOD weights and the other way around. \cref{tab:finetuning} shows that VaVit-L is able to exploit extra data and improve the performance on the target dataset, surpassing VaViT-B.}

\begin{table}[t]
    \centering
\setlength{\tabcolsep}{3pt}
    \resizebox{0.9\linewidth}{!}{%
    \renaud{\begin{tabular}{l|ccc|ccc}
    \toprule
            & SemKITTI && from WOD & WOD && from SemKITTI \\
    \midrule
    VaVit-L & 67.6     &\clap{$\Rightarrow$}& \textbf{69.3}     & 70.2 &\clap{$\Rightarrow$}& \textbf{71.1} \\
    \bottomrule
    \end{tabular}}}
    \caption{\renaud{\textbf{Exploiting a large capacity with multi-dataset finetuning.} We train a VaViT-L backbone on WOD, finetune it on SemKITTI, and evaluate (mIoU\%) on SemKITTI val set without TTA. Conversely, we finetune on WOD from SemKITTI weights.}}
    \label{tab:finetuning}
\end{table}

% ===============================================
\subsection{Robustness to global transformations and OOD}
\label{sec:tta}

\begin{table}[t]

\scriptsize
\centering
\ra{1.2}
\setlength{\tabcolsep}{4.5pt}
\begin{tabular}{lcccccc}
\toprule 
%\multirow{2}{*}{Method}
    & \multicolumn{2}{c}{nuScenes}
    & \multicolumn{2}{c}{SemKITTI}
    & \multicolumn{2}{c}{WOD}
\\
\cmidrule(lr){2-3}
\cmidrule(lr){4-5}
\cmidrule(lr){6-7}
Method
    & w/o TTA
    & w/~TTA
    & w/o TTA
    & w/~TTA
    & w/o TTA
    & w/~TTA
\\
\midrule
PTv3$^\dagger$
    & 78.9
    & 80.4
    & 66.2
    & \bf 68.8
    & 70.2
    & 71.7
\\
LitePT$^\dagger$
    & 80.7
    & \bf 82.2
    & \multicolumn{2}{c}{\textit{no result/code}}
    % & -
    & \bf 71.7
    & \bf 73.2
\\
\rowcolor{blue!15}VaViT-B*
    & \bf 81.3
    & 81.6
    & \bf 68.0
    & 68.7
    & 70.9
    & 71.4
\\
\bottomrule
\end{tabular}
\vspace{-7pt}
\caption{\textbf{Robustness to global transformations.} We report mIoU\% obtained on the validation sets of nuScenes, SemanticKITTI and WOD for PTV3, LitePT and VaViT-B*, with and without TTA. $^\dagger$ indicates results we reproduced.}
\label{tab:tta}
\end{table}

\paragraph{Use of TTAs.} A classical technique to improve performance is to use TTAs. A point cloud is typically transformed using global operations such as rotation around the vertical axis, scaling, and horizontal-plane flipping. While % , ideally, 
one would like a trained network to be robust to these transformations (used during the training phase), most networks are not. Ensembling predictions over %% multiple 
such transformations then boosts the performance. \renaudRoundOne{It provides a gain of} % This technique permits to gain 
about $2$ mIoU points for PTv3 and LitePT (%see the 
first two rows of \cref{tab:tta}).

In contrast, VaViT models are much more robust to such global transformations. The results in the last row of \cref{tab:tta} demonstrates this \renaudRoundOne{property} % behavior 
as the scores are nearly unchanged after applying TTAs with global transformations, \renaudRoundOne{although using 12 different augmentations with random parameters}: 
the transformations have been mostly learned by the networks.
We explain \renaudRoundOne{it} % this behavior 
by the use of global self-attentions in the ViT, instead of local attentions as in PTv3 and LitePT. From a global point of view, the structure of the point cloud is unchanged after applying global transformations, which the global attention mechanism is more likely to be robust to, as it has a view on the entire point cloud. The downside is that TTA with such transformations \renaudRoundOne{does not} % is not a technique that 
significantly boost performance of our VaViT models.

\paragraph{\renaud{Out-of-distribution (OOD) data.}}

\renaud{There is no inductive bias in VaViT, contrary to other approaches, and we study whether it has an impact on OOD data. To that end, we evaluate on the Robo3D benchmark \cite{kong2023robo3d}, made of datasets artificially corrupted from severe weather conditions, external disturbances, and sensor missing beams. With this benchmark, VaViT gets the lowest Corruption Error (mCE) and highest Resilience Rate (mRR) on nuScenes-C (82.1 and 83.9), and the best mCE on SemanticKITTI-C (92.9), showing the remarkable robustness of VaViT's architecture to OOD data. More details are provided in appendix.}

% ===============================================
%
% ===============================================
\section{Conclusion}

We have shown that, with a careful design of a tokenizer, a segmentation head and an augmentation, it is possible to use a plain ViT and compete with SOTA methods for the semantic segmentation of automotive 
lidar point clouds. 

We believe this work can open several doors, including easier multimodal architecture designs, \renaudRoundOne{including} mixing point clouds and images, and the direct application of self-supervised pretraining techniques for transformers, which have proven powerful in other domains \renaud{and could counteract scaling limitations for point clouds when the dataset is large but the labels reduced due to annotation costs.}

\paragraph*{Acknowledgments.}
We acknowledge EuroHPC Joint Undertaking for awarding the project ID EHPC-REG-2024R02-234 access to Karolina, Czech Republic. This work was also supported by the European Union’s Horizon Europe research and innovation programme under grant agreement No 101214398 (ELLIOT).
Gilles Puy contributed to this work while employed by Valeo; he accessed and processed all datasets on valeo.ai compute infrastructures; he is presently employed by Meta.

% ===============================================
%
% ===============================================

\clearpage

\maketitle

\newpage
\appendix

\setcounter{figure}{1}
\setcounter{table}{8}

\section{Implementation details}

Unless otherwise specified, we train our VaViT models with a batch size of 8, for 45 epochs and a weight decay of $0.003$, using AdamW \cite{adamw}. The learning rate schedule starts with a linear warmup during the first 4 epochs to reach $0.001$ and is followed by a cosine decrease to 0.  

We make use of standard augmentations: random rotations around the vertical axis, random flip of the horizontal plane, and random scaling in $[0.9, 1.1]$. On SemanticKITTI, we use the CutMix augmentation \cite{RPVNet,2dpass} as available in the official implementation of \cite{waffleiron}. 

We use stochastic depth \cite{stocdepth} in the ViT backbone with a drop probability of~0.3, \renaud{except for SemanticKITTI where we use a drop probability of 0.5}. 

For \pmixplus, 
unless otherwise mentioned, we mix three different point clouds using square pillars of side size $S\,{=}\,$7.5\,m. The pillar size $s$ for tokenization is set to 
0.5\,m for all datasets.

\section{Details about our adaptation of FlatFormer}
\label{sec:flatformer}

FlatFormer \cite{flatformer} was designed for object detection using a BEV representation of point clouds.
It uses a flattened window attention mechanism and alternates flattening directions within groups.
As mentioned in the main paper, we adapt it to semantic segmentation for comparison purposes and refer to it as Flatformer-S.

We directly replace our ViT by Flatformer's transformer architecture.
That is, the proposed Flatformer-S uses our tokenizer and our segmentation head.
In our implementation of Flatformer-S, we adapt the dimensions and number of blocks to match the ViT-S architecture.
We use a feature dimension of 384, 3 groups of 4 flattened attention blocks (12 transformer blocks) and 6 heads.
The BEV resolution is the same as for VaViT for a fair comparison.

\section{\renaud{Latency and memory consumption.}}

\renaud{The main paper provides in Tab.\,2 the latency and memory consumption of VaViT, compared to PTv3 and LitePT, for a single forward pass, thus without TTA. In \cref{tab:runningtimes_tta} of this appendix, we complement Tab.\,2 with runtime and memory requirements when TTAs are used.}

\renaud{Please note that, as the results of VaViT with TTAs are not significantly larger than without TTA (Tab.\,8), the practical regime of VaViT is not to use any TTA. In contrast, other methods require the TTA trick to be or come close to SOTA. In this case, as can be seen in \cref{tab:runningtimes_tta}, the latency of VaViT is at least one order of magnitude smaller than the latency of other methods.}

\begin{table*}[t]
\small
\centering
\setlength{\tabcolsep}{3pt}
\resizebox{0.99\linewidth}{!}{\renaud{\begin{tabular}{l|ccc|ccc|ccc}
\toprule
& \multicolumn{3}{c|}{Latency without TTA} & \multicolumn{3}{c|}{Latency with TTA} & \multicolumn{3}{c}{GPU memory consumption}\\
         & nuScenes      & SemanticKITTI & WOD             & nuScenes           & SemanticKITTI      & WOD                & nuScenes & SemanticKITTI & WOD \\ 
\midrule
PTv3     & 59\,ms (50\,ms) & 111\,ms (87\,ms) & 142\,ms (82\,ms) & 96: 5.7\,s (4.8\,s) & 69: 7.6\,s (6.0\,s) & 24: 3.4\,s (2.0\,s) & 0.6 GB & 1.5 GB & 1.0 GB \\
LitePT-S & 46\,ms (35\,ms) & \textit{no code} & 104\,ms (48\,ms) & 92: 4.2\,s (3.2\,s) & \textit{no code} & 24: 2.5\,s (1.2\,s) & 0.5 GB & \textit{no code} & 0.9 GB \\
VaViT-B  & 17\,ms (17\,ms) &  70\,ms (70\,ms) & 109\,ms (109\,ms) & 12: 0.2\,s (0.2\,s) & 12: 0.8\,s (0.8\,s) & 12: 1.4\,s (1.4\,s) & 0.7 GB & 1.7 GB & 1.9 GB \\
\bottomrule
\end{tabular}}}
\caption{\renaud{\textbf{Evaluation of the efficacy of semantic segmentation.} All runtime measurements are made on the same Nvidia H100 GPU. They exclude I/Os and are made after a warm-up stage. The number in parentheses is the time spent only in the GPU, thus excluding I/Os, dataloader preprocessing and CPU. Both values are the same for VaViT as it only uses the GPU, contrary to PTv3 and LitePT that resort to sparse 3D convolutions and thus also use of the CPU. ``latency without TTA'' measures the time of a single forward pass, with batch size 1. The number before ``:'' is the actual number of TTA passes (including chunking). For PTv3 and LitePT, the augmentations and their number differ depending on the dataset. Additionally, this number of passes may vary due to the randomized grid downsampling made for each augmentation.}}
\label{tab:runningtimes_tta}
\end{table*}

\begin{table*}[t]
\small
\newcommand*\rotext{\multicolumn{1}{R{65}{1em}}}
\centering
\ra{1.3}
\setlength{\tabcolsep}{2pt}
\renewcommand{\arraystretch}{1}
\renaud{\begin{tabular}{l | c | c c c c c c c c c c c c c c c c c c c}
\toprule 
Seed
    & \rotext{\bf mIoU\%}
    & \rotext{car}
    & \rotext{bicycle}
    & \rotext{motorcycle}
    & \rotext{truck}
    & \rotext{other-vehicle}
    & \rotext{person}
    & \rotext{bicyclist}
    & \rotext{motorcyclist}
    & \rotext{road}
    & \rotext{parking}
    & \rotext{sidewalk}
    & \rotext{other-ground}
    & \rotext{building}
    & \rotext{fence}
    & \rotext{vegetation}
    & \rotext{trunk}
    & \rotext{terrain}
    & \rotext{pole}
    & \rotext{traffic-sign}
\\
\midrule
      0 (paper) & 67.6 & 95.7 & 56.5 & 74.9 & 95.4 & 52.1 & 81.4 & 94.5 & 0.1 & 95.6 & 51.1 & 83.8 & 0.1 & 90.6 & 61.0 & 88.1 & 72.0 & 73.5 & 65.9 & 51.6 \\
      2026 & 67.9 & 95.9 & 56.6 & 78.4 & 95.5 & 55.4 & 80.5 & 93.9 & 0.0 & 95.7 & 50.3 & 83.9 & 0.4 & 91.0 & 63.4 & 88.0 & 72.4 & 73.1 & 65.4 & 49.5 \\  
    1990 & 68.1 & 95.7 & 55.1 & 79.0 & 95.6 & 56.8 & 81.6 & 94.6 & 0.0 & 95.9 & 52.3 & 83.9 & 0.7 & 91.1 & 64.0 & 88.2 & 72.9  & 73.8 &  65.7  & 48.9 \\
\midrule
    mean & 67.8 &	95.8 & 56.1 & 77.4&	95.5 &	54.8 &	81.2 &	94.3 &	0.0	& 95.7 &	51.2 &	83.9 & 0.4 & 90.9 &	62.8 & 88.1 & 72.4 & 73.7	& 65.5 & 48.9 \\
\midrule
    std dev &  0.4	& 0.1 &	0.8 &	2.2	& 0.1 &	2.4	& 0.6 &	0.4 &	0.1 &	0.2	& 1.0 & 0.1	& 0.3 & 0.3 & 1.6 & 0.1 & 0.5	& 0.5 & 0.2 & 0.7 \\
      \bottomrule
    \end{tabular}}
\caption{\renaud{Performance of VaViT-B on SemanticKITTI validation set, without TTA, for 3 trainings with different randomization seeds.}}
\label{tab:perfsk3runs}
\label{hop}
\makeatletter
% \show\r@hop@cref
\makeatother
\end{table*}
\begin{table*}[t!]
\newcommand*\rotext{\multicolumn{1}{R{65}{1em}}}
\centering
\small
\ra{1.3}
\newcommand\UA{\,$\uparrow$}%
\newcommand\DA{\,$\downarrow$}%
\setlength{\tabcolsep}{3pt}
\renaud{%
%\resizebox{0.99\linewidth}{!}{
\begin{tabular}{l|c|cccccccc}
\toprule
Dataset          & \relax{mRR\UA} & \rotext{Fog} & \rotext{Wet} & \rotext{Snow} & \rotext{Move} & \rotext{Beam} & \rotext{Cross} & \rotext{Echo} & \rotext{Sensor} \\
\midrule
nuScenes-C      & 83.88 & 93.37 & 98.86 & 90.88 & 68.93 & 88.68 & 64.83 & 82.54 & 64.83 \\
SemanticKITTI-C & 80.86 & 81.65 & 91.90 & 89.03 & 36.20 & 93.01 & 78.82 & 89.00 & 87.24 \\
\bottomrule
\end{tabular}%
}
%}%
%\vspace{-7pt}
\caption{\renaud{\textbf{Classwise Resilience Rate (RR) of VaViT-B on Robo3D \cite{kong2023robo3d}} for corruptions fog, wet ground, snow, motion blur, beam missing,
crosstalk, incomplete echo, and cross-sensor.}}
\label{tab:oodrobustclass}
\vspace{-3mm}
\end{table*}

\begin{table}[t]
\centering
\newcommand\UA{\,$\uparrow$}%
\newcommand\DA{\,$\downarrow$}%
\renaud{%
\resizebox{0.99\linewidth}{!}{\begin{tabular}{l@{\,}r|cc|cc}
\toprule
&& \multicolumn{2}{c|}{\small nuScenes-C} & \multicolumn{2}{c}{\small SemanticKITTI-C} \\
Method          && mCE\DA & mRR\UA & mCE\DA & mRR\UA \\
\midrule
WaffleIron      & \cite{waffleiron} & 106.73 & 72.78 & 109.54 & 72.18 \\
Cylinder3D-TSC  & \cite{Cylinder3d} & 105.56 & 78.08 & 103.13 & \textbf{83.90} \\
2DPASS          & \cite{2dpass}     &  98.56 & 75.24 & 106.14 & 77.50 \\
SPVCNN-34       & \cite{spvnas}     &  97.45 & 75.10 & \uline{99.16} & 82.01 \\
MinkUNet-34     & \cite{minknet}    &  96.37 & 75.08 & 100.61 & 80.22 \\
KPConv          & \cite{kpconv}     &    --- &   --- &  99.54 & \uline{82.90} \\
RPVNet          & \cite{RPVNet}     &    --- &   --- & 111.74 & 73.86 \\
GFNet           & \cite{gfnet}      & \uline{92.55} & \uline{83.31} & 108.68 & 77.92 \\
\midrule
VaViT-B           & \llap{(ours)}     & \textbf{82.06} & \textbf{83.88} & \textbf{92.84} & 80.86 \\
\bottomrule
\end{tabular}%
}}%
% \vspace{-7pt}
\caption{\renaud{\textbf{Evaluation of OOD robustness on Robo3D \cite{kong2023robo3d}.}}}
\label{tab:oodrobust}
\vspace{-5mm}
\end{table}

\section{\renaud{Variance study on SemanticKITTI}}

\renaud{\cref{tab:perfsk3runs} provides results for three different trainings of VaViT-B on SemanticKITTI, evaluated on the validation set. It provides an estimation of the standard deviation for both the mIoU (estimated at 0.3\%) and the classwise IoUs.}

\renaud{Unsurprisingly, classes with high variance correspond to rare or notoriously difficult classes, whose performance is bad (bicycle, other vehicle, parking, traffic sign). For extremely difficult classes (motorcyclist, other ground), the result is so bad (close to 0 mIoU) that there is on the contrary almost no variance. Other methods show similar bad performance for these classes (see Tab.\,1 in the paper) as well as similar variance when retraining.}

\section{\renaud{Robustness to OOD data}}

\renaud{We evaluated the robustness of VaViT-B to OOD data on the Robo3D benchmark \cite{kong2023robo3d}, using the synthetically-corrupted datasets nuScenes-C and SemanticKITTI-C. In \cref{tab:oodrobust}, we compare to the best and second best methods in \cite{kong2023robo3d} regarding metrics mCE (mean Corruption Error), mRR (mean Resilience Rate) and mIoU, as well as to methods in Tab.~1 which have already been evaluated on Robo3D. VaViT gets the best mCE and mRR on nuScenes-C and the best mCE on SemanticKITTI-C, showing an excellent robustness on average to the eight Robo3D corruptions types. Individual Resilience Rates for each corruption (fog, wet ground, snow, motion blur, beam missing, crosstalk, incomplete echo and cross-sensor) are given in \cref{tab:oodrobustclass}.}

\section{Attention maps}
\label{sec:attention}

\begin{figure*}
\centering
\includegraphics[width=.75\linewidth]{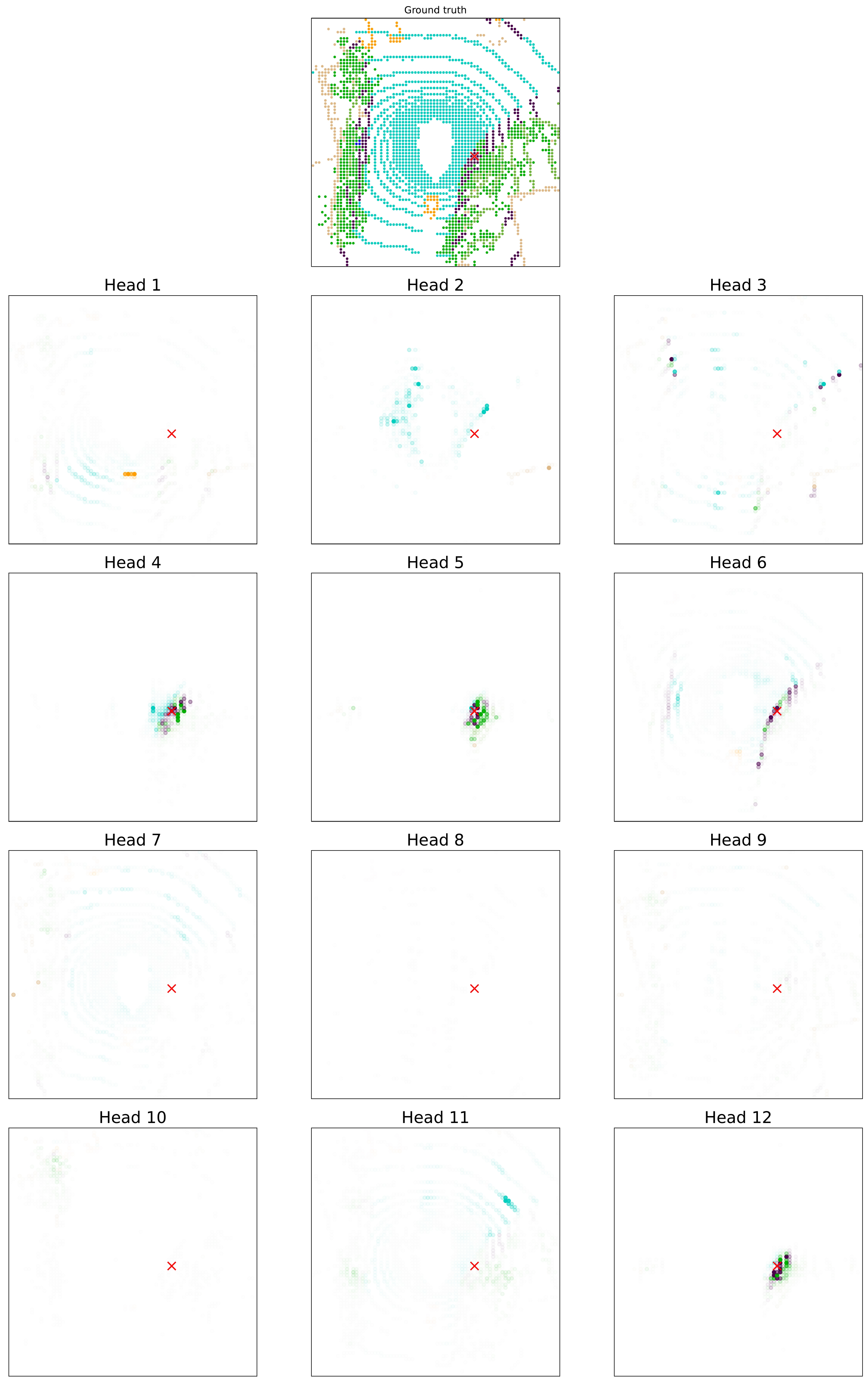}
\caption{\textbf{Attention maps} for each head at the \textbf{1$^{\rm st}$} (first) layer of our VaViT-B model trained on \textbf{nuScenes}. The query point, denoted by a red cross, is located on the \textbf{sidewalk}. Ground truth in BEV is presented at the top. Subsequent maps have a transparency scaled by the attention weight between the query point and the keys; points with zero attention are fully transparent.}
\label{fig:attention_nuscenes_sidewalk_first}
\end{figure*}

\begin{figure*}
\centering
\includegraphics[width=.75\linewidth]{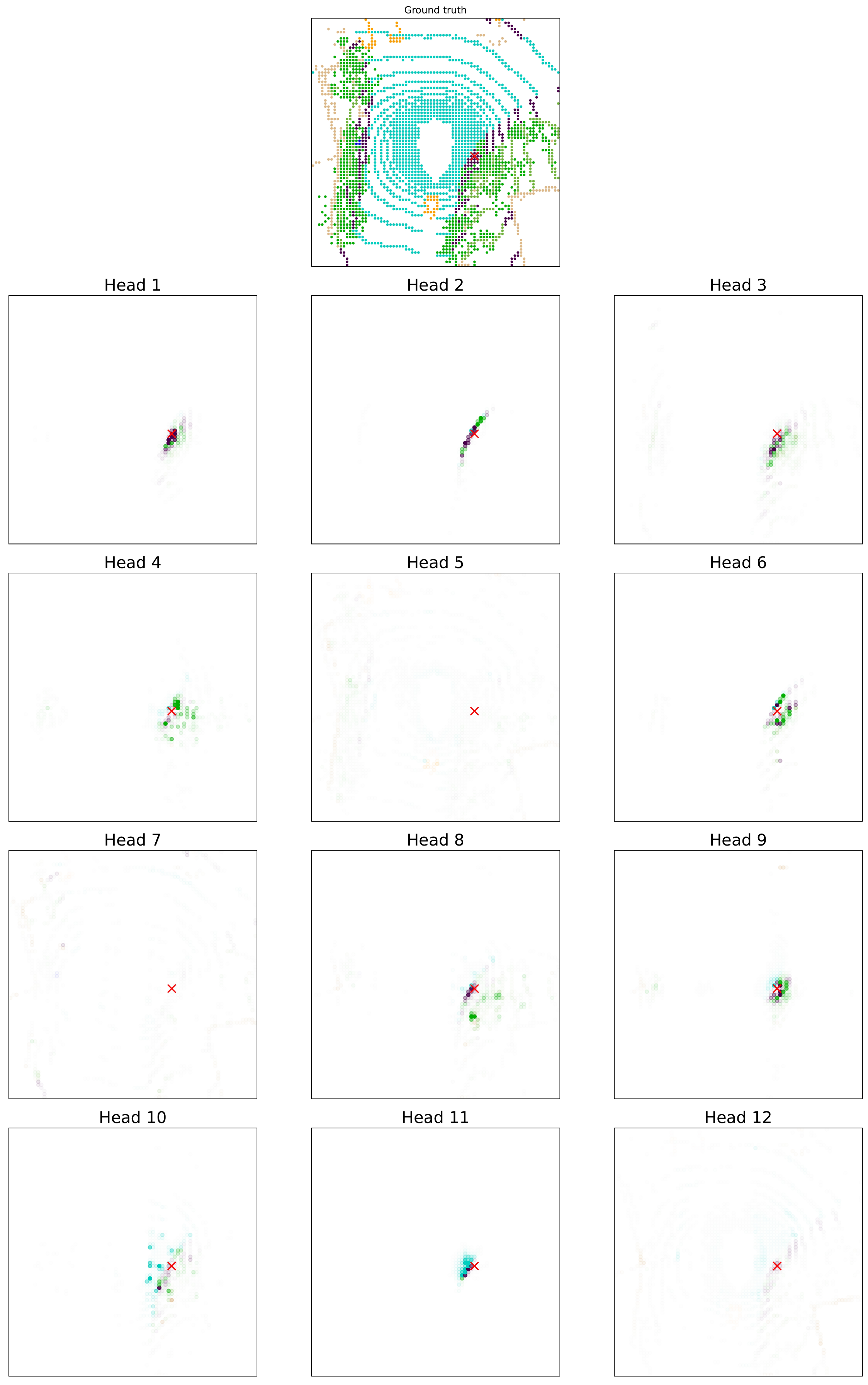}
\caption{\textbf{Attention maps} for each head at the \textbf{6$^{\rm th}$} (middle) layer of our VaViT-B model trained on \textbf{nuScenes}. The query point, denoted by a red cross, is located on the \textbf{sidewalk}. Ground truth in BEV is presented at the top. Subsequent maps have a transparency scaled by the attention weight between the query point and the keys; points with zero attention are fully transparent.}
\label{fig:attention_nuscenes_sidewalk_mid}
\end{figure*}

\begin{figure*}
\centering
\includegraphics[width=.75\linewidth]{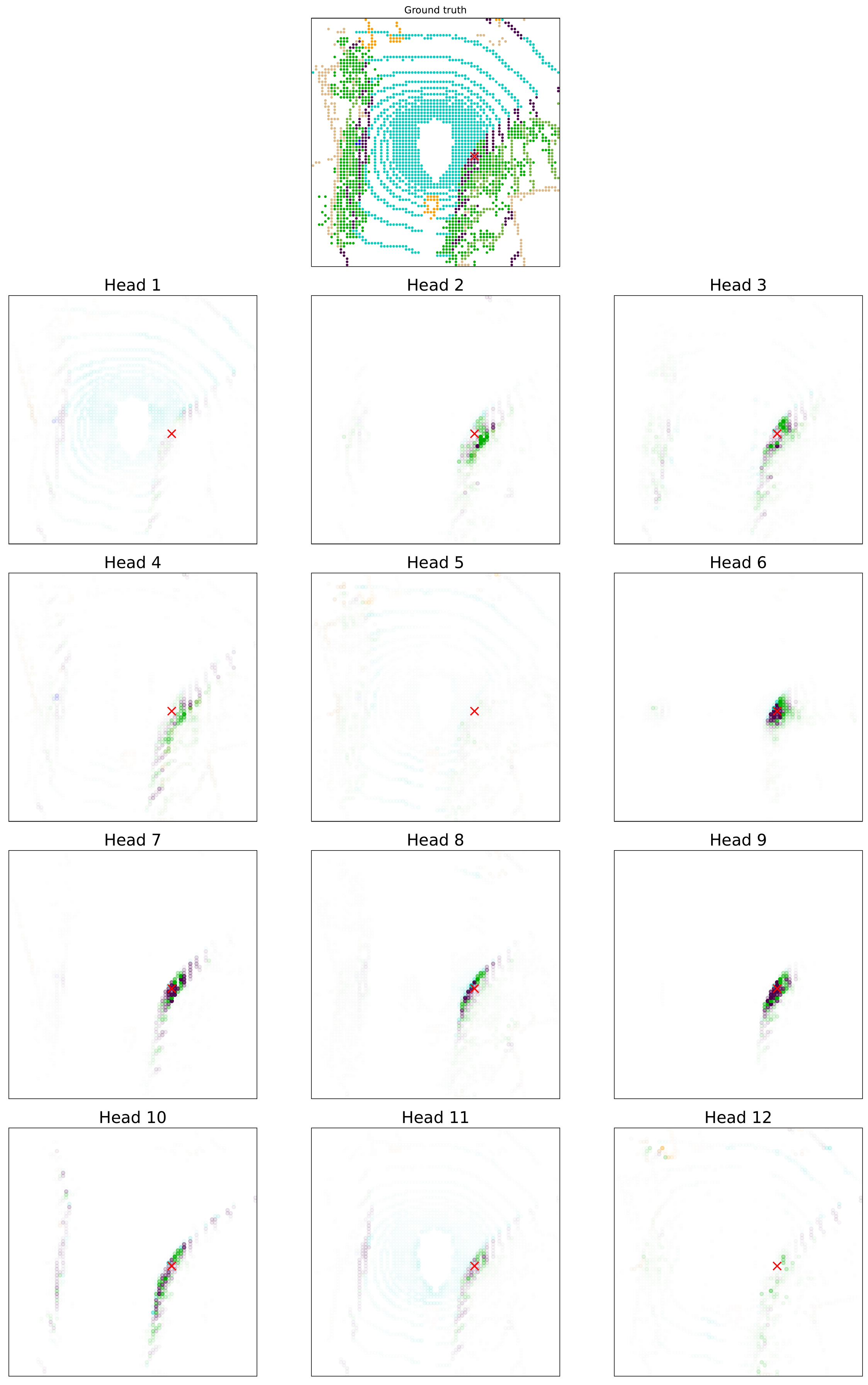}
\caption{\textbf{Attention maps} for each head at the \textbf{12$^{\rm th}$} (final) layer of our VaViT-B model trained on \textbf{nuScenes}. The query point, denoted by a red cross, is located on the \textbf{sidewalk}. Ground truth in BEV is presented at the top. Subsequent maps have a transparency scaled by the attention weight between the query point and the keys; points with zero attention are fully transparent.}
\label{fig:attention_nuscenes_sidewalk_last}
\end{figure*}

\begin{figure*}
\centering
\includegraphics[width=.75\linewidth]{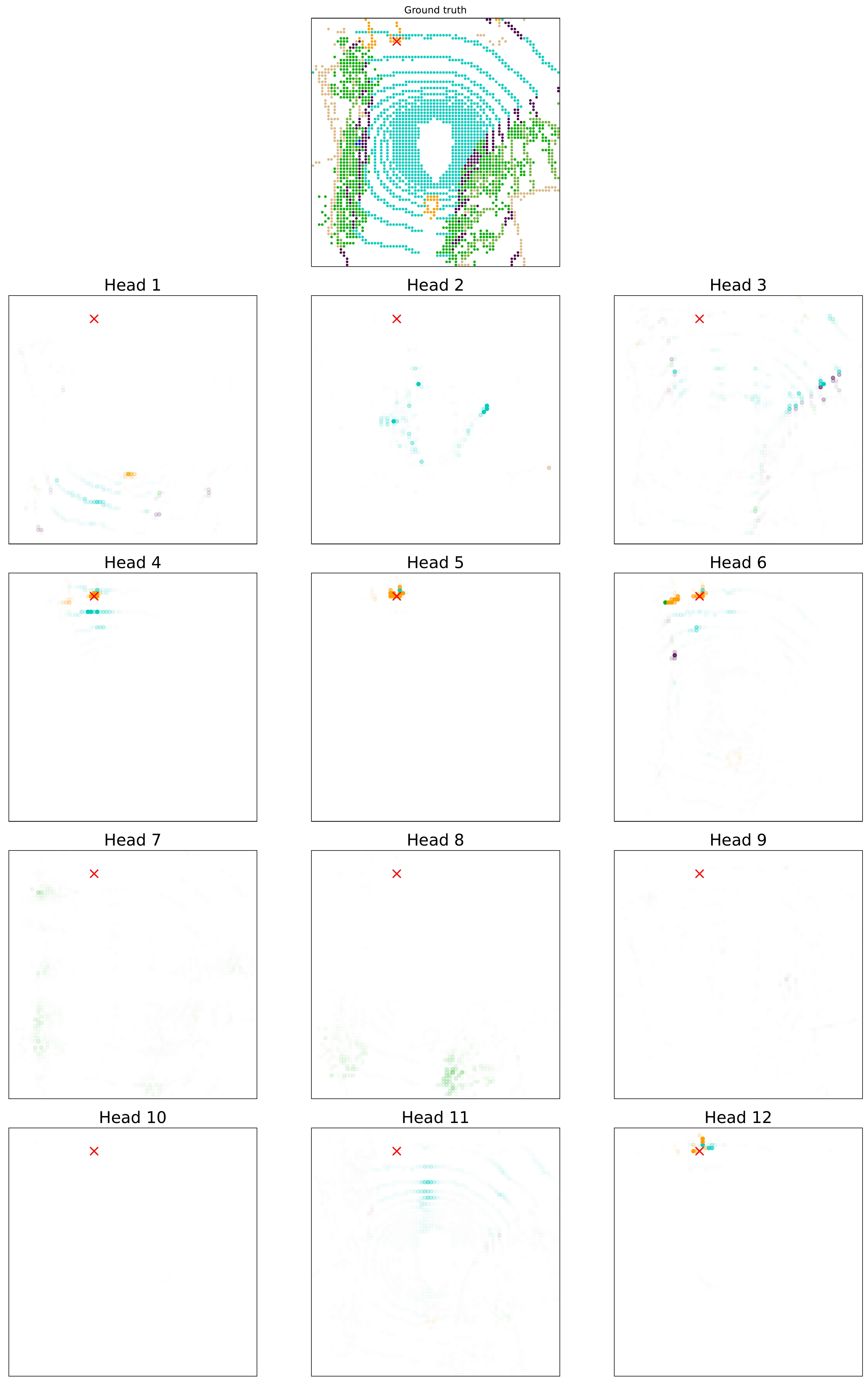}
\caption{\textbf{Attention maps} for each head at the \textbf{1$^{\rm st}$} (first) layer of our VaViT-B model trained on \textbf{nuScenes}. The query point, denoted by a red cross, is located on a \textbf{car}. Ground truth in BEV is presented at the top. Subsequent maps have a transparency scaled by the attention weight between the query point and the keys; points with zero attention are fully transparent.}
\label{fig:attention_nuscenes_car_first}
\end{figure*}

\begin{figure*}
\centering
\includegraphics[width=.75\linewidth]{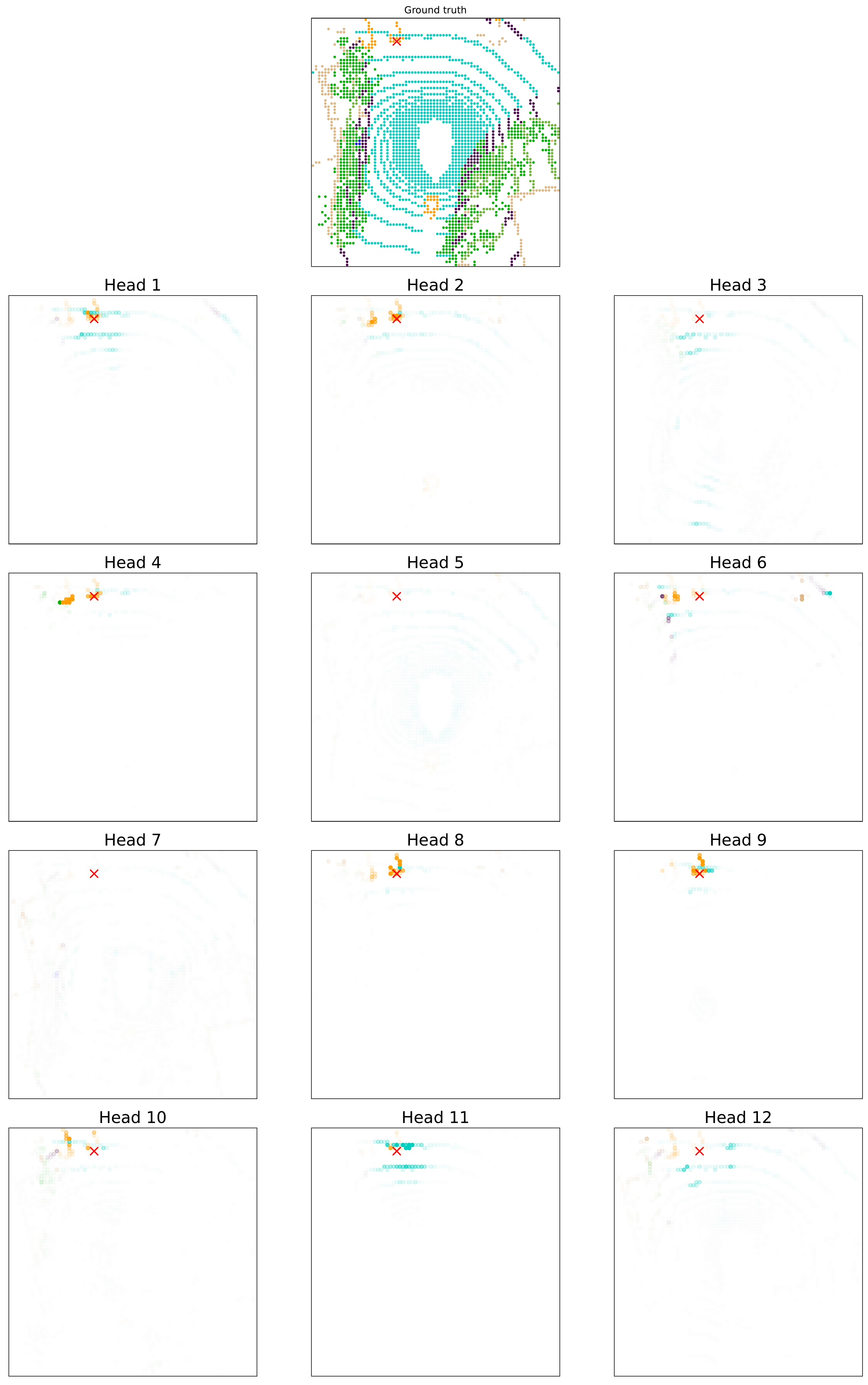}
\caption{\textbf{Attention maps} for each head at the \textbf{6$^{\rm th}$} (middle) layer of our VaViT-B model trained on \textbf{nuScenes}. The query point, denoted by a red cross, is located on a \textbf{car}. Ground truth in BEV is presented at the top. Subsequent maps have a transparency scaled by the attention weight between the query point and the keys; points with zero attention are fully transparent.}
\label{fig:attention_nuscenes_car_mid}
\end{figure*}

\begin{figure*}
\centering
\includegraphics[width=.75\linewidth]{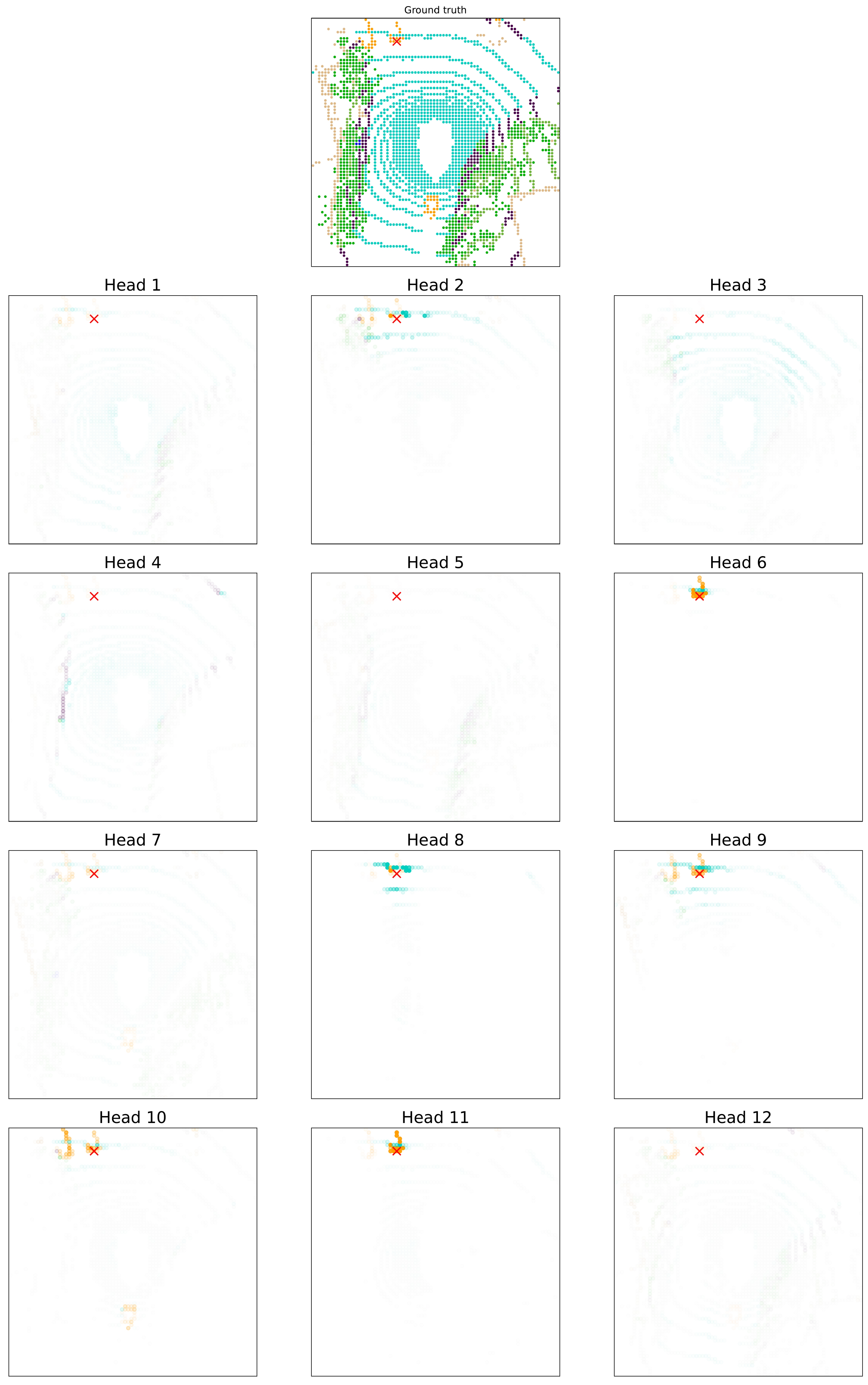}
\caption{\textbf{Attention maps} for each head at the \textbf{12$^{\rm th}$} (last) layer of our VaViT-B model trained on \textbf{nuScenes}. The query point, denoted by a red cross, is located on a \textbf{car}. Ground truth in BEV is presented at the top. Subsequent maps have a transparency scaled by the attention weight between the query point and the keys; points with zero attention are fully transparent.}
\label{fig:attention_nuscenes_car_last}
\end{figure*}

\begin{figure*}
\centering
\includegraphics[width=.75\linewidth]{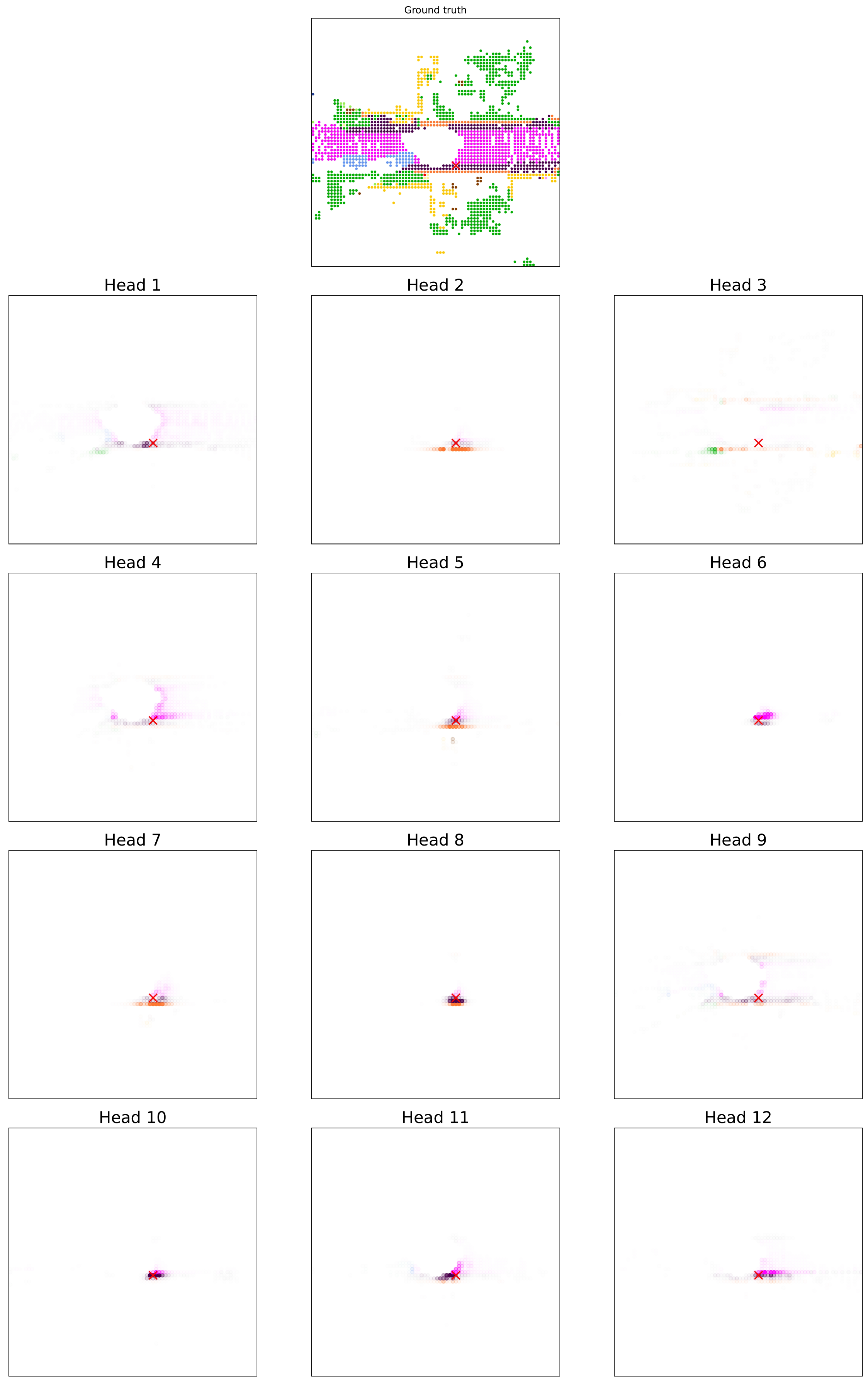}
\caption{\textbf{Attention maps} for each head at the \textbf{1$^{\rm st}$} (first) layer of our VaViT-B model trained on \textbf{SemanticKITTI}. The query point, denoted by a red cross, is located on the \textbf{sidewalk}. Ground truth in BEV is presented at the top. Subsequent maps have a transparency scaled by the attention weight between the query point and the keys; points with zero attention are fully transparent.}
\label{fig:attention_kitti_sidewalk_first}
\end{figure*}

\begin{figure*}
\centering
\includegraphics[width=.75\linewidth]{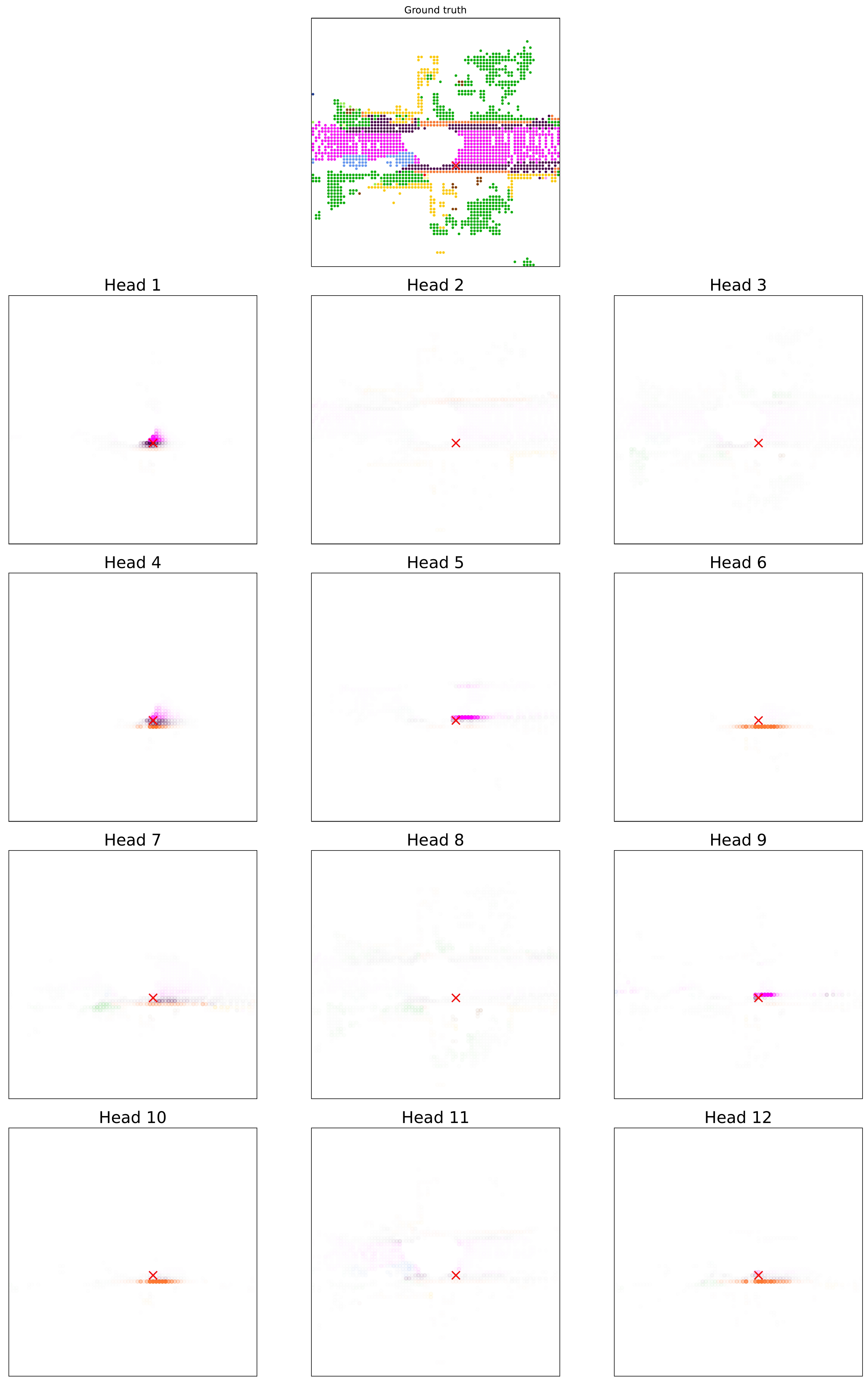}
\caption{\textbf{Attention maps} for each head at the \textbf{6$^{\rm th}$} (middle) layer of our VaViT-B model trained on \textbf{SemanticKITTI}. The query point, denoted by a red cross, is located on the \textbf{sidewalk}. Ground truth in BEV is presented at the top. Subsequent maps have a transparency scaled by the attention weight between the query point and the keys; points with zero attention are fully transparent.}
\label{fig:attention_kitti_sidewalk_mid}
\end{figure*}

\begin{figure*}
\centering
\includegraphics[width=.75\linewidth]{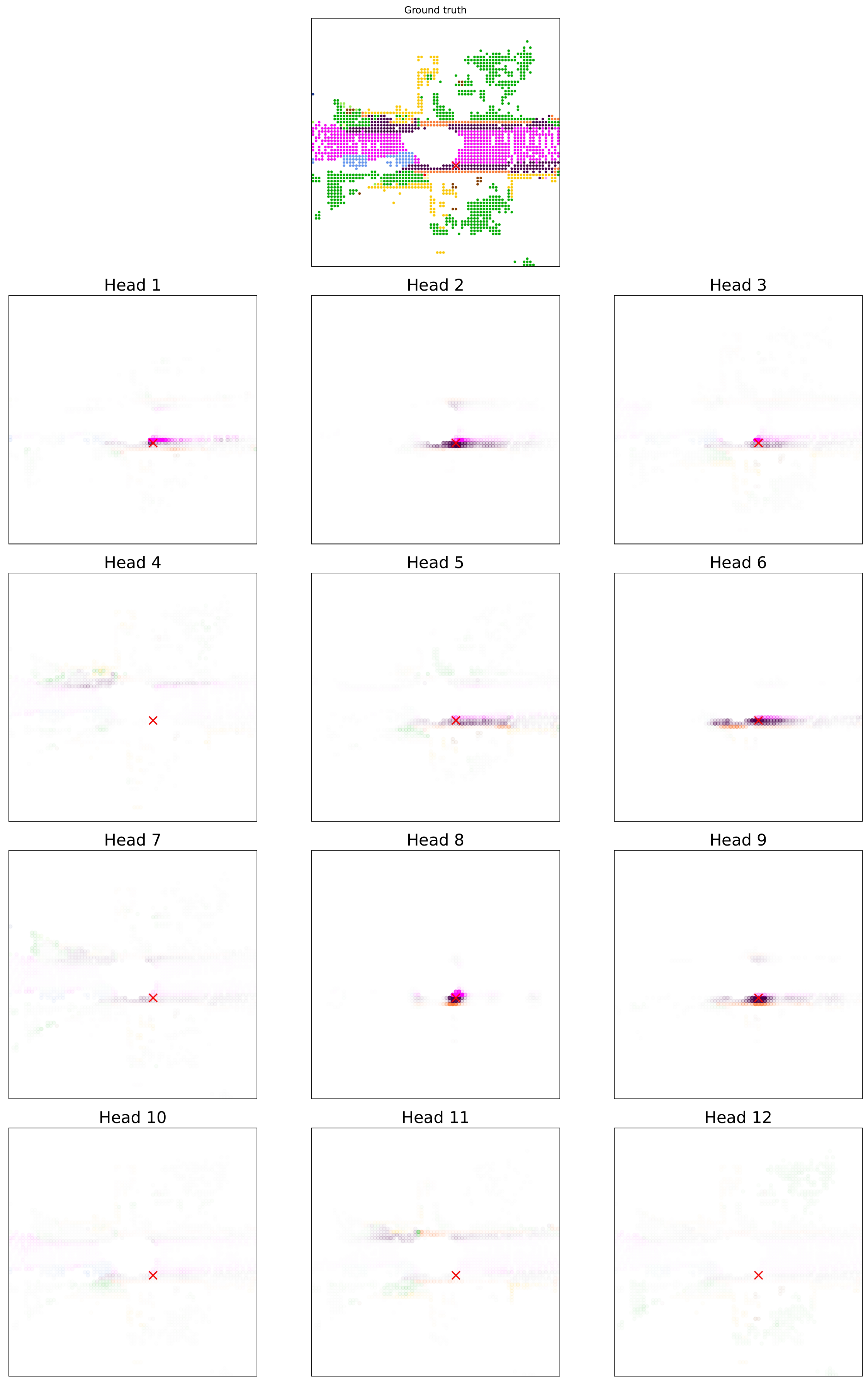}
\caption{\textbf{Attention maps} for each head at the \textbf{12$^{\rm th}$} (last) layer of our VaViT-B model trained on \textbf{SemanticKITTI}. The query point, denoted by a red cross, is located on the \textbf{sidewalk}. Ground truth in BEV is presented at the top. Subsequent maps have a transparency scaled by the attention weight between the query point and the keys; points with zero attention are fully transparent.}
\label{fig:attention_kitti_sidewalk_last}
\end{figure*}

\begin{figure*}
\centering
\includegraphics[width=.75\linewidth]{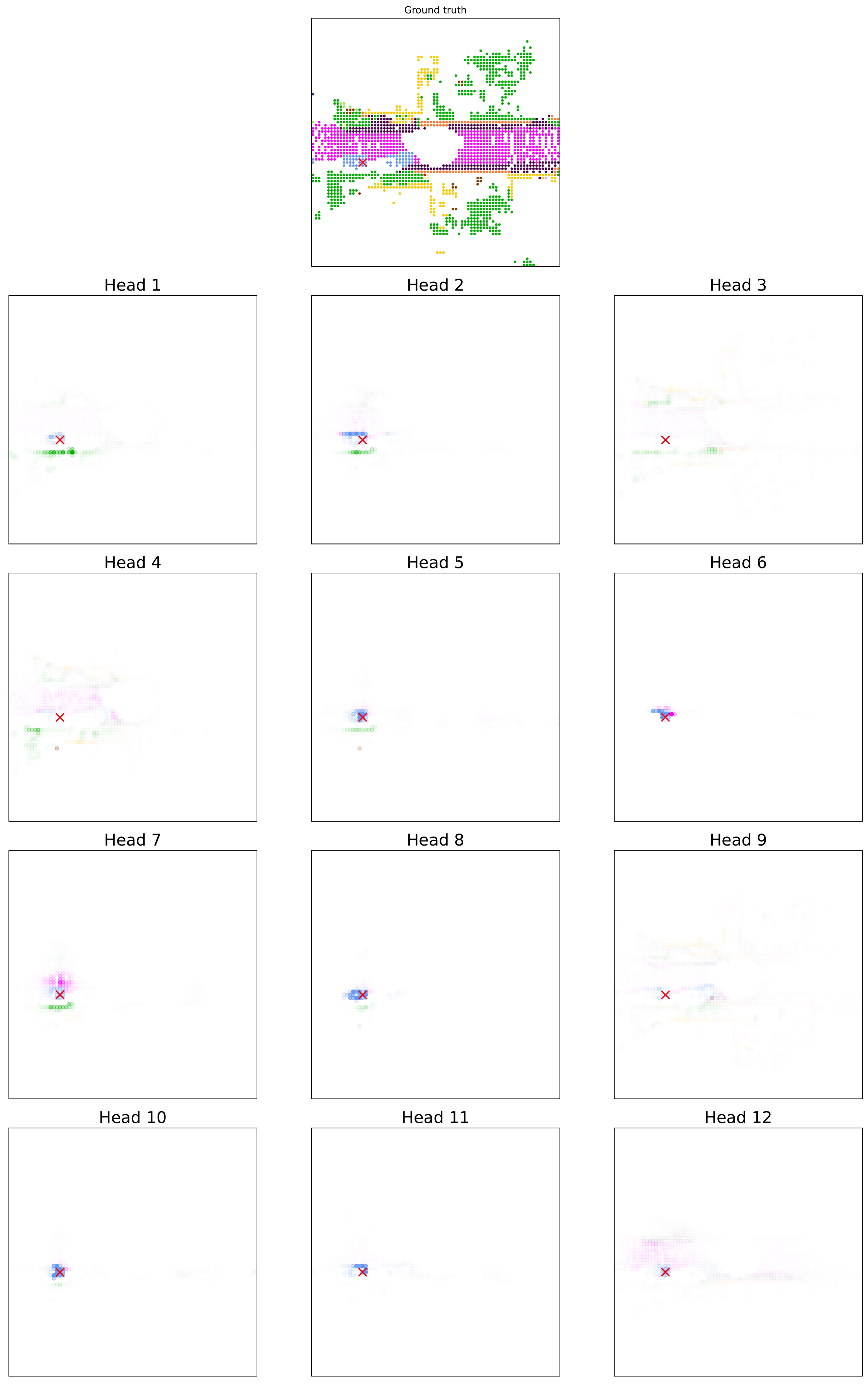}
\caption{\textbf{Attention maps} for each head at the \textbf{1$^{\rm st}$} (first) layer of our VaViT-B model trained on \textbf{SemanticKITTI}. The query point, denoted by a red cross, is located on a \textbf{car}. Ground truth in BEV is presented at the top. Subsequent maps have a transparency scaled by the attention weight between the query point and the keys; points with zero attention are fully transparent.}
\label{fig:attention_kitti_car_first}
\end{figure*}

\begin{figure*}
\centering
\includegraphics[width=.75\linewidth]{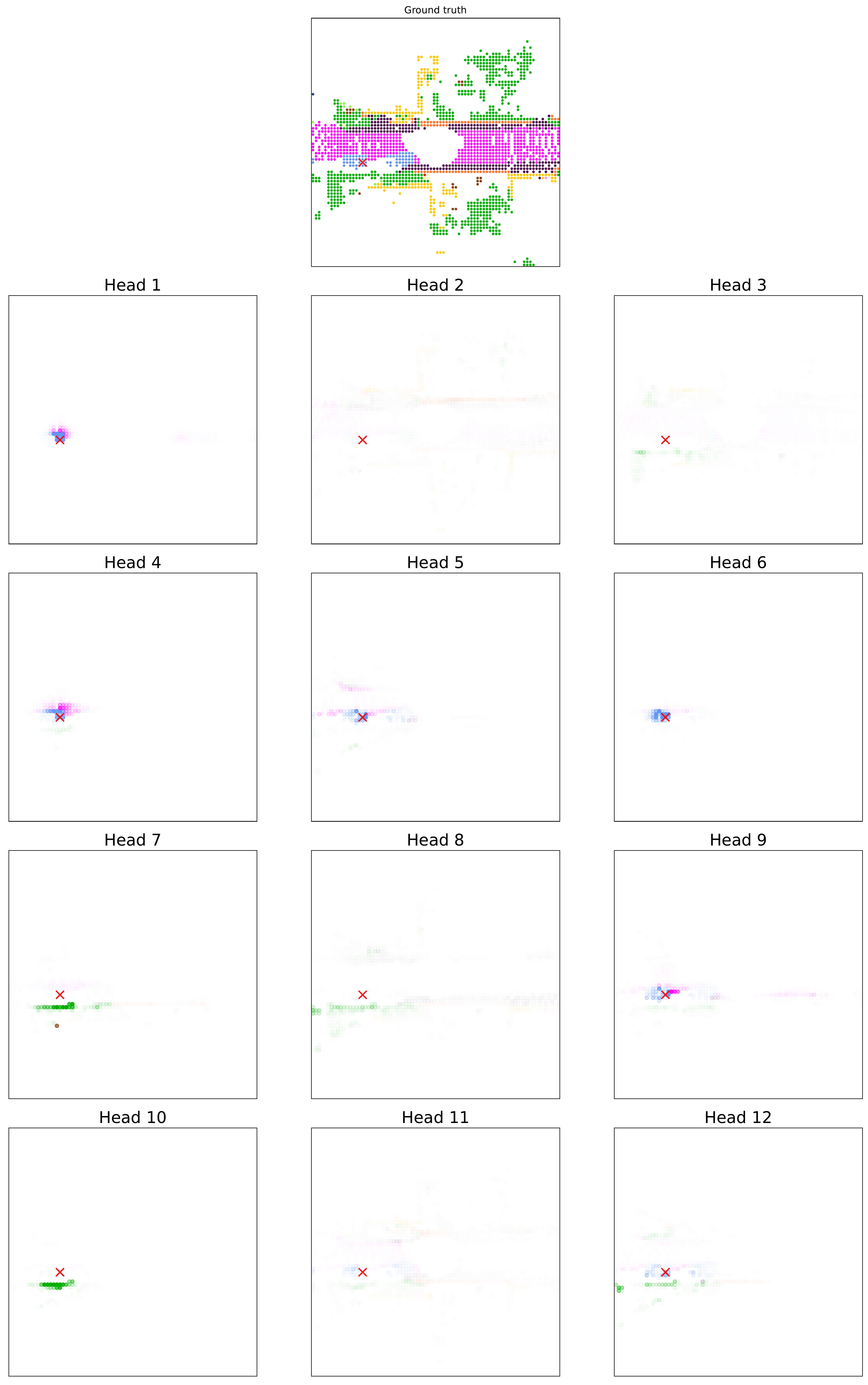}
\caption{\textbf{Attention maps} for each head at the \textbf{6$^{\rm th}$} (middle) layer of our VaViT-B model trained on \textbf{SemanticKITTI}. The query point, denoted by a red cross, is located on a \textbf{car}. Ground truth in BEV is presented at the top. Subsequent maps have a transparency scaled by the attention weight between the query point and the keys; points with zero attention are fully transparent.}
\label{fig:attention_kitti_car_mid}
\end{figure*}

\begin{figure*}
\centering
\includegraphics[width=.75\linewidth]{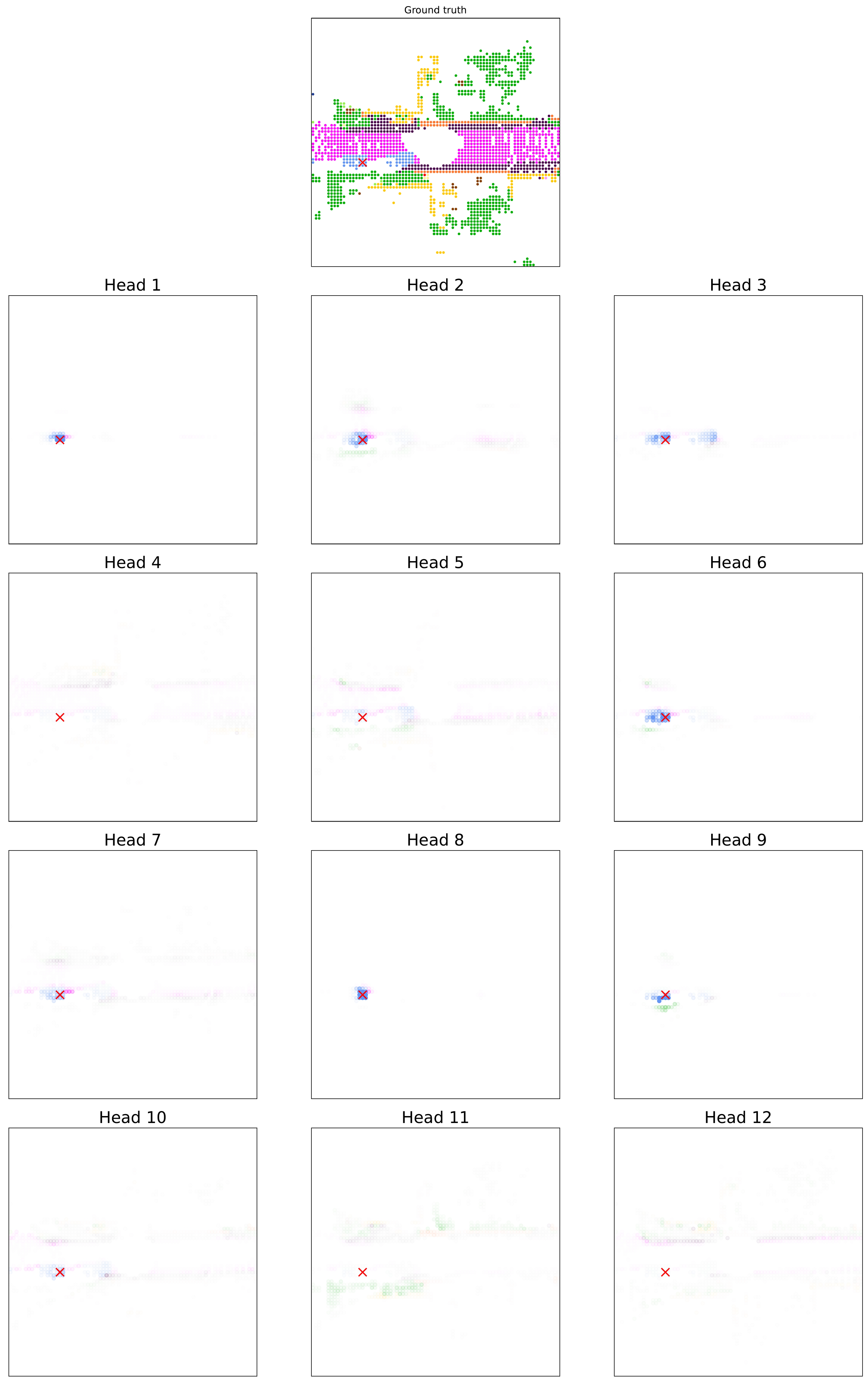}
\caption{\textbf{Attention maps} for each head at the \textbf{12$^{\rm th}$} (last) layer of our VaViT-B model trained on \textbf{SemanticKITTI}. The query point, denoted by a red cross, is located on a \textbf{car}. Ground truth in BEV is presented at the top. Subsequent maps have a transparency scaled by the attention weight between the query point and the keys; points with zero attention are fully transparent.}
\label{fig:attention_kitti_car_last}
\end{figure*}

\begin{figure*}
\centering
\includegraphics[width=.75\linewidth]{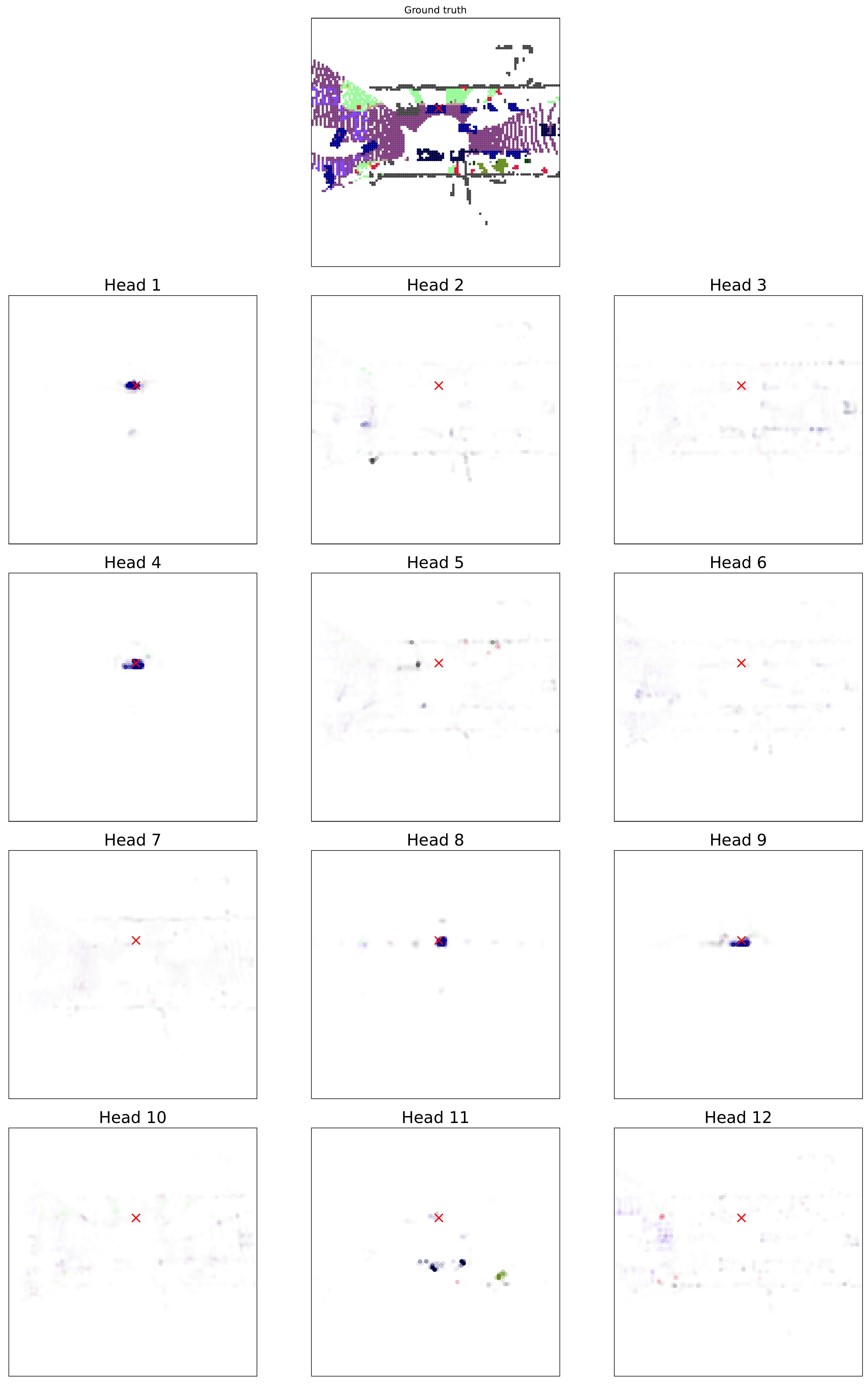}
\caption{\textbf{Attention maps} for each head at the \textbf{1$^{\rm st}$} (first) layer of our VaViT-B model trained on \textbf{WOD}. The query point, denoted by a red cross, is located on a \textbf{car}. Ground truth in BEV is presented at the top. Subsequent maps have a transparency scaled by the attention weight between the query point and the keys; points with zero attention are fully transparent.}
\label{fig:attention_waymo_car_first}
\end{figure*}

\begin{figure*}
\centering
\includegraphics[width=.75\linewidth]{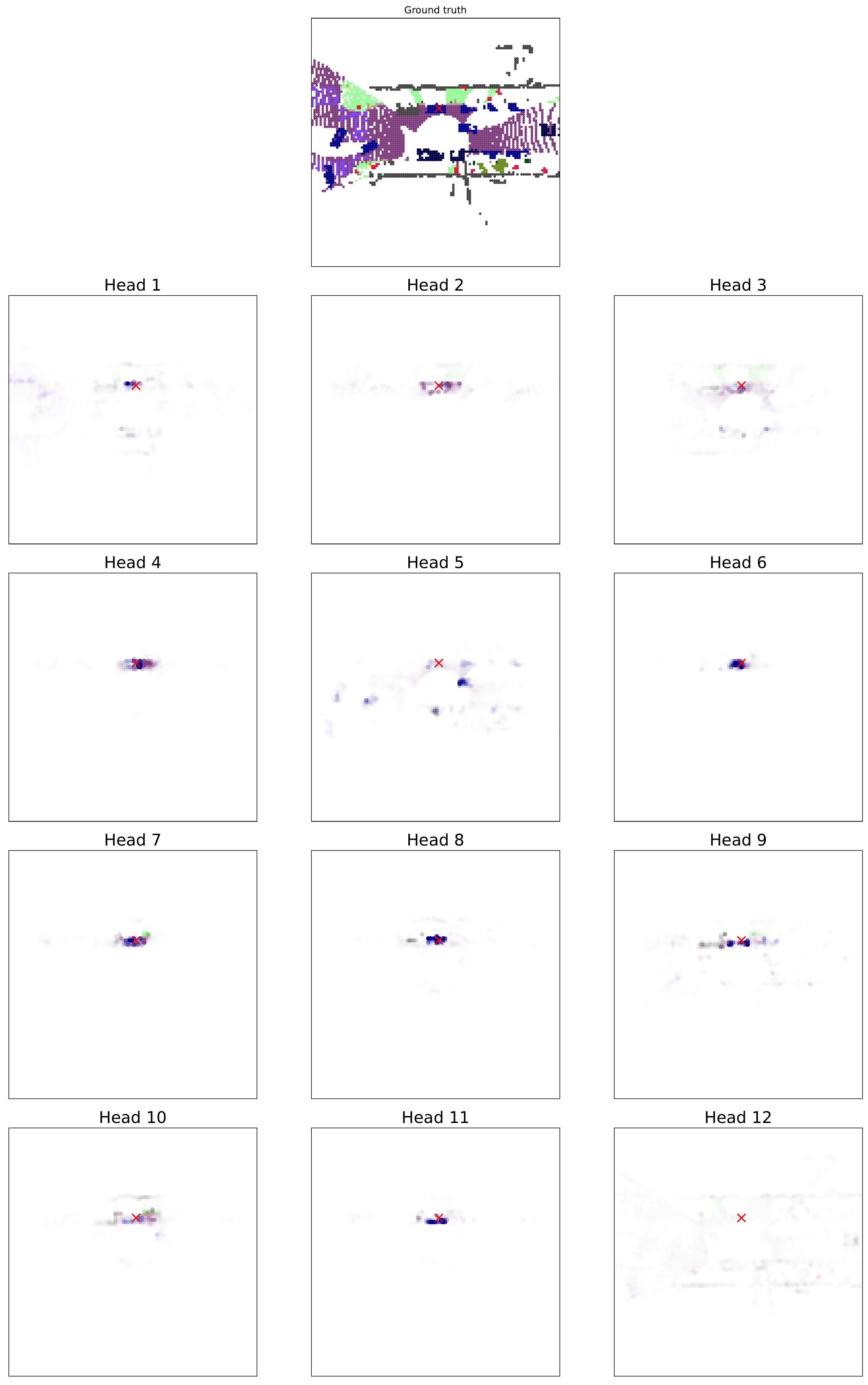}
\caption{\textbf{Attention maps} for each head at the \textbf{6$^{\rm th}$} (middle) layer of our VaViT-B model trained on \textbf{WOD}. The query point, denoted by a red cross, is located on a \textbf{car}. Ground truth in BEV is presented at the top. Subsequent maps have a transparency scaled by the attention weight between the query point and the keys; points with zero attention are fully transparent.}
\label{fig:attention_waymo_car_mid}
\end{figure*}

\begin{figure*}
\centering
\includegraphics[width=.75\linewidth]{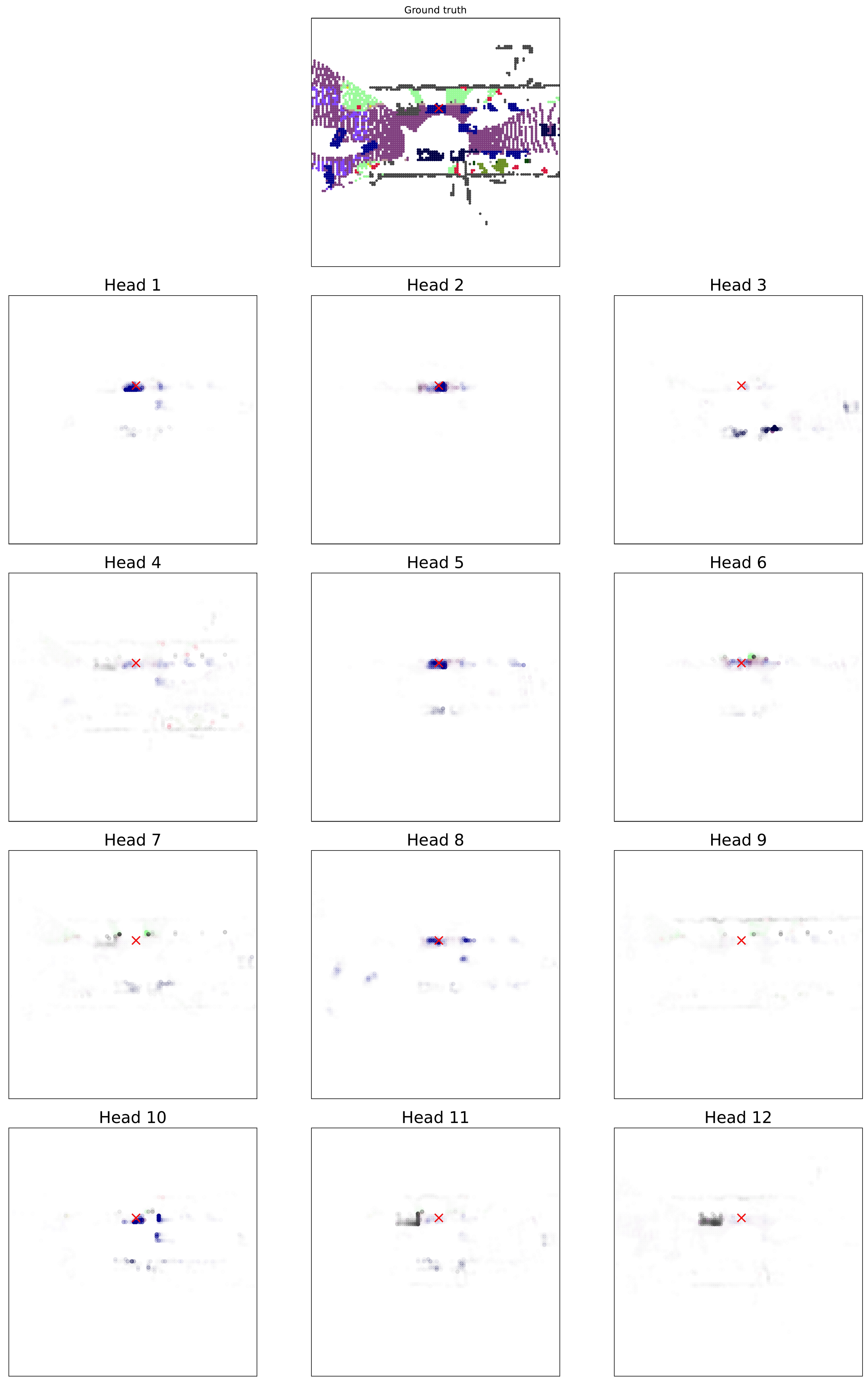}
\caption{\textbf{Attention maps} for each head at the \textbf{12$^{\rm th}$} (last) layer of our VaViT-B model trained on \textbf{WOD}. The query point, denoted by a red cross, is located on a \textbf{car}. Ground truth in BEV is presented at the top. Subsequent maps have a transparency scaled by the attention weight between the query point and the keys; points with zero attention are fully transparent.}
\label{fig:attention_waymo_car_last}
\end{figure*}

\begin{figure*}
\centering
\includegraphics[width=.75\linewidth]{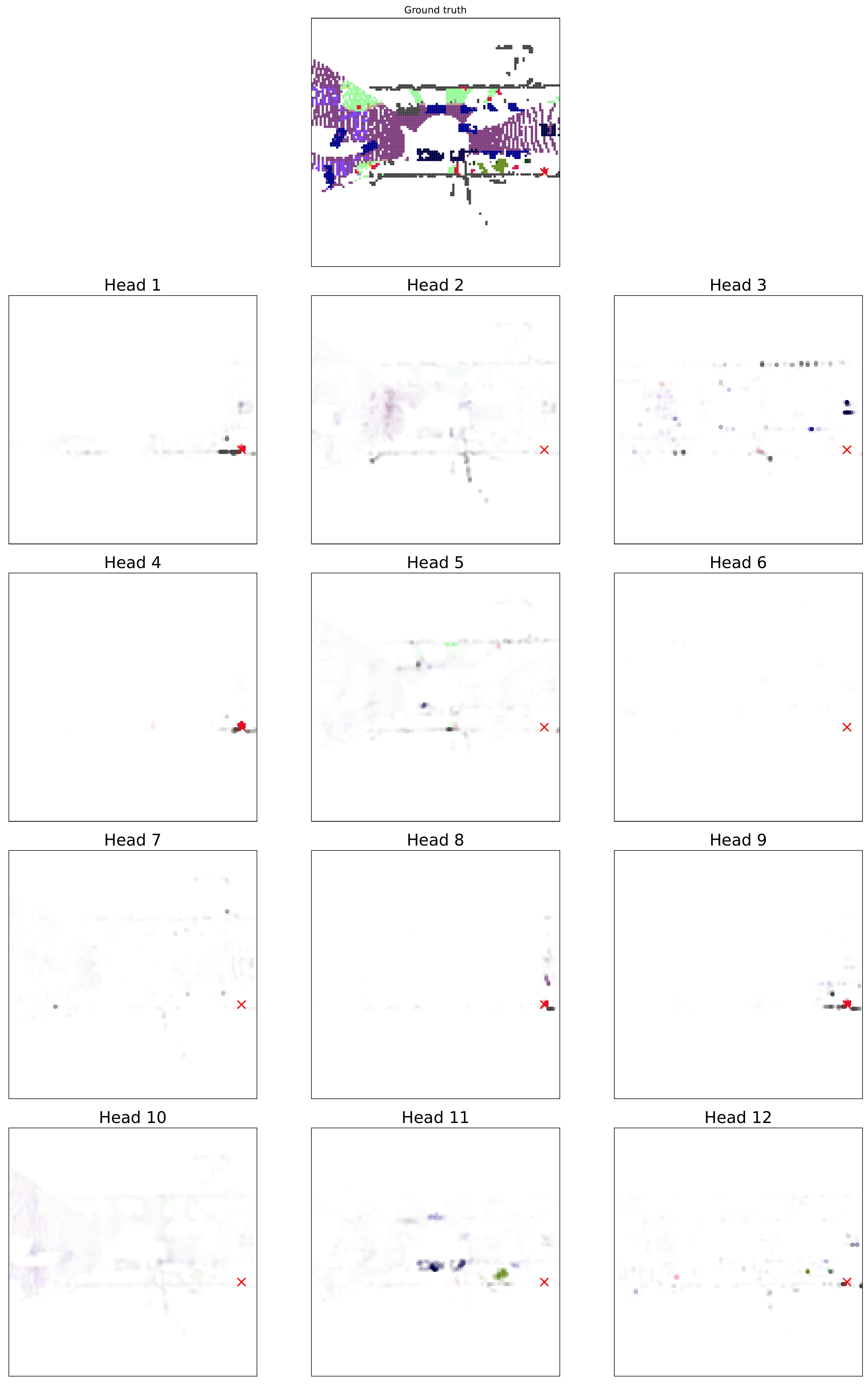}
\caption{\textbf{Attention maps} for each head at the \textbf{1$^{\rm st}$} (first) layer of our VaViT-B model trained on \textbf{WOD}. The query point, denoted by a red cross, is located on a \textbf{pedestrian}. Ground truth in BEV is presented at the top. Subsequent maps have a transparency scaled by the attention weight between the query point and the keys; points with zero attention are fully transparent.}
\label{fig:attention_waymo_pedestrian_first}
\end{figure*}

\begin{figure*}
\centering
\includegraphics[width=.75\linewidth]{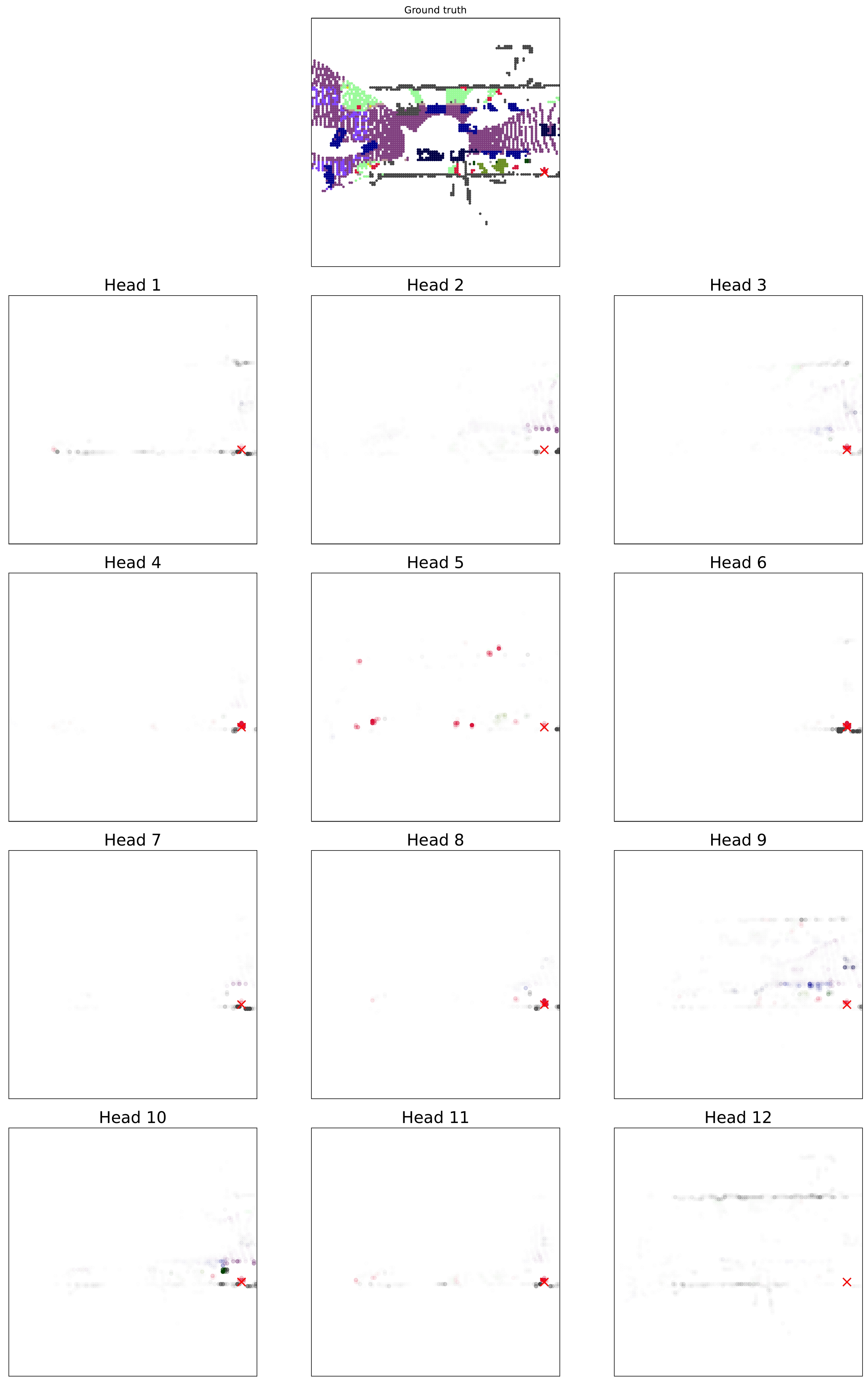}
\caption{\textbf{Attention maps} for each head at the \textbf{6$^{\rm th}$} (middle) layer of our VaViT-B model trained on \textbf{WOD}. The query point, denoted by a red cross, is located on a \textbf{pedestrian}. Ground truth in BEV is presented at the top. Subsequent maps have a transparency scaled by the attention weight between the query point and the keys; points with zero attention are fully transparent.}
\label{fig:attention_waymo_pedestrian_mid}
\end{figure*}

\begin{figure*}
\centering
\includegraphics[width=.75\linewidth]{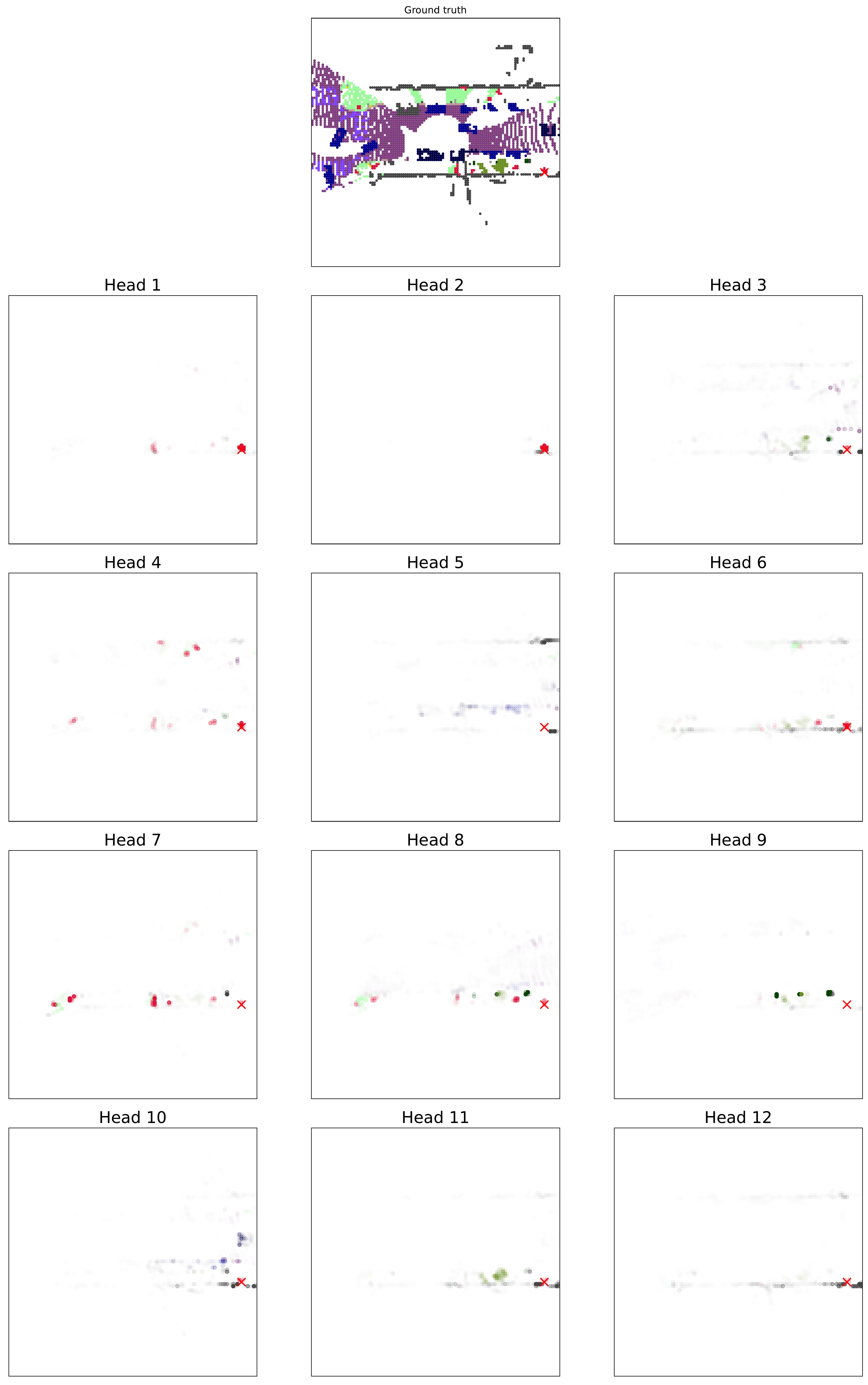}
\caption{\textbf{Attention maps} for each head at the \textbf{12$^{\rm th}$} (last) layer of our VaViT-B model trained on \textbf{WOD}. The query point, denoted by a red cross, is located on a \textbf{pedestrian}. Ground truth in BEV is presented at the top. Subsequent maps have a transparency scaled by the attention weight between the query point and the keys; points with zero attention are fully transparent.}
\label{fig:attention_waymo_pedestrian_last}
\end{figure*}

Last, We present attention maps at the first, middle and last layers of our VaViT-B models when trained on 
nuScenes \cite{nuscenes} in \cref{fig:attention_nuscenes_sidewalk_first,fig:attention_nuscenes_sidewalk_mid,fig:attention_nuscenes_sidewalk_last,fig:attention_nuscenes_car_first,fig:attention_nuscenes_car_mid,fig:attention_nuscenes_car_last},
SemanticKITTI \cite{semkitti} in
\cref{fig:attention_kitti_sidewalk_first,fig:attention_kitti_sidewalk_mid,fig:attention_kitti_sidewalk_last,fig:attention_kitti_car_first,fig:attention_kitti_car_mid,fig:attention_kitti_car_last},
and WOD \cite{waymo} in
\cref{fig:attention_waymo_car_first,fig:attention_waymo_car_mid,fig:attention_waymo_car_last,fig:attention_waymo_pedestrian_first,fig:attention_waymo_pedestrian_mid,fig:attention_waymo_pedestrian_last}. The attention maps are presented at Layer 1, 6 and 12 of the VaViT-B backbones. We use query points on car and sidewalk for nuScenes and SemanticKITTI, and car and pedestrian on WOD.

In general, as we go deeper in the network, the attention tends to concentrate on regions containing points of similar semantics. 
It is interesting to see, e.g., that 
when querying a point on a pedestrian, the attention maps in Head~5 of Layer~6 (\cref{fig:attention_waymo_pedestrian_mid}) and Head~4 of Layer~12 highlight the other pedestrians in the scene (\cref{fig:attention_waymo_pedestrian_last}).

\newpage
% ======================================
%
% ======================================

\clearpage
{
    \bibliographystyle{ieeenat_fullname}
    \bibliography{main}  % .bib
}

\end{document}